\documentclass[journal]{IEEEtran}
\usepackage[utf8]{inputenc}
\usepackage[T1]{fontenc}
\usepackage{multicol}
\usepackage{textcomp}
\usepackage{amsmath}
\usepackage{amssymb}
\usepackage{filecontents,lipsum}
\usepackage[noadjust]{cite}
\usepackage{footnote}
\usepackage{cprotect}
\usepackage{fvextra}
\usepackage[dvipsnames]{xcolor}
\usepackage{graphicx}
\usepackage{subfig}
\usepackage[ruled,vlined]{algorithm2e}
\usepackage{multicol,tabularx,capt-of}
\usepackage{multirow}
\usepackage{float}
\usepackage{stackengine}
\usepackage[hidelinks]{hyperref}
\usepackage{hyperref}
\usepackage{array}

\begin{document}

\title{Road Segmentation for Remote Sensing Images using Adversarial Spatial Pyramid Networks }

\author{Pourya~Shamsolmoali, ~\IEEEmembership{Member,~IEEE,}
              Masoumeh~Zareapoor, Huiyu~Zhou, Ruili Wang, and~Jie~Yang
              \thanks{Manuscript received September 13,  2019; revised February 29, 2020; accepted August 10, 2020.~~ This work is supported in part by NSFC under Grant 61806125, 61977046. ~ Corresponding author: Jie Yang (jieyang@sjtu.edu.cn).}
\thanks{P.~Shamsolmoali, M.~Zareapoor, J.~Yang are with the Institute of Image Processing and Pattern Recognition, Shanghai Jiao Tong University, Shanghai 200240, China.}
\thanks{H. Zhou is with the School of Informatics, University of Leicester, Leicester LE1 7RH, United Kingdom.}
\thanks{R. Wang is with the School of Logistics and Transportation, Central South University of Forestry and Technology, China, and the School of Natural and Computational Sciences, Massey University, Auckland, New Zealand.}
}

\maketitle

\begin{abstract}
\textcolor{blue}{To read the paper please go to IEEE Transactions on Geoscience and Remote Sensing on IEEE Xplore.} Road extraction in remote sensing images is of great importance for a wide range of applications. Because of the complex background, and high density, most of the existing methods fail to accurately extract a road network that appears correct and complete. Moreover, they suffer from either insufficient training data or high costs of manual annotation. To address these problems, we introduce a new model to apply structured domain adaption for synthetic image generation and road segmentation. We incorporate a feature pyramid network into generative adversarial networks to minimize the difference between the source and target domains. A generator is learned to produce quality synthetic images, and the discriminator attempts to distinguish them. We also propose a feature pyramid network that improves the performance of the proposed model by extracting effective features from all the layers of the network for describing different scales’ objects. Indeed, a novel scale-wise architecture is introduced to learn from the multi-level feature maps and improve the semantics of the features. For optimization, the model is trained by a joint reconstruction loss function, which minimizes the difference between the fake images and the real ones. A wide range of experiments on three datasets prove the superior performance of the proposed approach in terms of accuracy and efficiency. In particular, our model achieves state-of-the-art $78.86$ IOU on the Massachusetts dataset with $14.89$M parameters and $86.78$B FLOPs, with $4\times$ fewer FLOPs but higher accuracy ($+3.47\%$ IOU) than the top performer among state-of-the-art approaches used in the evaluation.
\end{abstract}


\IEEEpeerreviewmaketitle

\section{Introduction}
\IEEEPARstart{C}{urrently} road segmentation in remote sensing (RS) images has become one of the crucial tasks in many urban applications such as traffic management, urban planning and road monitoring. Meanwhile, it is tremendously time-consuming to manually label roads from the high-resolution images. Unsupervised models, which are based on the predefined features, achieved low accuracy and failed on heterogeneous regions. However, supervised deep learning models, have achieved high performance in most of computer vision tasks, such as object detection \cite{zhao2019m2det, li2019nested, henry2018road}, semantic segmentation \cite{chen2019residual, yu2018semantic, li2018road}, and skeleton extraction \cite{liu2018roadnet}. With the improvement of convolution neural networks, road detection from RS images tends to be an efficient and effective process.\par
Actually, the road network detection contains two subtasks: road edge detection and road surface segmentation that meets several problems: semantic segmentation, and object extraction. Generally, buildings, cars and trees along the roads create shadows or can lead to occlusions that cause heterogeneous regions on roads. Hence, road detection is a challenging task. Hence, road detection is a challenging task. However, the current methods fail to tackle the above-mentioned issues, in which the benchmark datasets are usually selected from the urban areas. 
With the great success of convolutional neural networks and fully connected networks, several architectures have been proposed for RS road detection and segmentation \cite{henry2018road, li2018road}. On the other hand, most of the state-of-the-art road detection methods are fully supervised, demanding massive labeled training data. Data augmentation is a technique that is generally used in the state-of-the-art methods for increasing the number of the training data. Nevertheless, if the distribution between the training and the testing datasets is changed, the performance can still be unsatisfactory. Another approach is using generative adversarial networks (GANs) to produce synthetic images~\cite{bermudez2019synthesis}. However, synthetic data is generally more problematic than the realistic data. By using simulation tools, a huge amount of synthetic images for training a neural network can be produced \cite{huang2013unified, xu2018satellite}. Nevertheless, the recent works \cite{hong2018conditional, sun2019learning} reveal that, still there is a gap between the distribution of the real and that of the synthetic data. 

It has been observed that using deep features resulted in minimizing the cross-domain distribution discrepancy, but did not remove it \cite{teng2019classifier}. Transfer learning is a machine learning method in which a developed model for a task is reused for a model on the other task, which is widely used in RS image classification \cite{dong2019sig}. Transfer learning algorithms, unlike traditional machine learning algorithms, intend to build models that are applicable for different domains or tasks. Domain adaptation is a transfer learning technique, in which the standard domain assumption does not hold, and the learned models from one or more domains may apply for the same task in another domain. In domain adaptation, training is conducted in the source domain, and the test samples are taken from the target domain \cite{ganin2016domain}. This paper targets at training a road segmentation model for real RS images and parsing them without using any manual annotation. This problem is worthy to be investigated due to three reasons: 1) Road detection in RS images has become an attractive theme in both industry and academia, while segmentation is one of the most obligatory procedures in understanding high density and complicated RS images; 2) A large amount of high-resolution and unseen annotated imagery are required to well train a deep neural network for RS road segmentation. In comparison with object detection and classification in RS data, it is an extremely laborious and time-consuming task for a professional to annotate pixel-wise training data for image segmentation. For instance, a synthetic image annotation on average only takes a few seconds. In contrast, single image manual annotation may take more than 1.5 hours \cite{cordts2016cityscapes}; 3) this process requires dedicated equipment and time to gather images which cover a large diverse urban landscape in different areas with different lighting set-ups. 
Theoretically and practically, it is interesting and worthy to develop a method to conduct automated road segmentation on RS images without manual labeling. \par
One of the most powerful methods for creating synthetic images is GANs \cite{cordts2016cityscapes}. GANs generally consist of two networks in which the first network, the generator, learns to produce synthesis images that are barely distinguishable from the genuine images. On the other hand, the other network (discriminator) is trained to distinguish the real images from the generated ones. Simultaneously, the generator and the discriminator are trained to minimize the adversarial loss. Most of the present GANs methods generally focus on improving the loss function \cite{salimans2018improving}, designing new architectures or tricks \cite{odena2017conditional}, or creating new training schemes like the gradual and progressive training \cite{shamsolmoali2019g}. So far, limited works have been reported on the design of neural networks architecture for synthetic image generation \cite{hong2018conditional, sun2019learning}. 

The intuitive idea of this paper is to explore new architectures that minimize the feature space gap between the source and target domains, which can improve the performance of adversarial learning for generating realistic synthetic remote sensing images for road segmentation. The network is trained to transform the feature maps of source domain images to those of target domain images while preserving the semantic information of the features. We propose a generic and light-weight network architecture that easily can be integrated into the generator. We also introduce an efficient spatial pyramid network to extract multiscale and global contextual information. The proposed Adversarial Spatial Pyramid Network (ASPN) does not require matching image pairs from the corresponding domains that practically do not exist. To the best of our knowledge, this is the first work that integrates spatial feature pyramid networks with the adversarial domain adaptation for the RS road segmentation. Compared to the existing approaches, the proposed unsupervised domain adaptation architecture has the following advantages:

\begin{itemize}
\item{Network Architecture: Previous approaches \cite{dong2019sig, teng2019classifier} are generally based on two domains through an intermediate feature space and thereby implicitly take the same decision function for both domains. ASPN uses the multiscale feature maps with a conditional generator and residual learning; all the modules of the proposed architecture are placed in a single network and have an end-to-end training strategy. A discriminator is also used \cite{sun2019learning} to manage adversarial learning.}
\item{In the generator, we propose an efficient pyramid network architecture to extract effective pyramid features of multiple scales to overcome the current models' limitations. In the proposed pyramid network, firstly, the multiscale features are extracted and fused as the base features; we next feed them into an optimized U-shape network (OUN) \cite{zhao2019m2det} and Feature Modules Concatenation (FMC) to produce additional descriptive and multiscale features. We also use a module called Scale-wise feature Concatenate (SFC) which is similar to an inception network \cite{yu2018semantic} to merge the features in a wide range of scales. Finally, the feature maps with the same size are collected to form the last feature pyramid. The code will be made available on {\tt \url{https://github.com/pshams55/ASPN}}.}
\end{itemize}

\section{RELATED WORK}
Here we discuss recent works on road segmentation, domain adaptation and feature pyramid networks. 
\subsection{Road Segmentation}
The urban development led to significant growth of transportation networks. Such infrastructure requires frequent updates of road maps. Previous road detection and segmentation models were based on local features handcrafted by domain experts \cite{sghaier2015road}. Liu et al. \cite{liu2018roadnet} proposed a multitask pixel wise end-to-end CNN, to simultaneously predict road surfaces, edges, and centerlines. The model learns multiscale and multilevel features and is trained in a specially designed cascaded network, which can deal with the roads in various scenes and scales that deals with several issues in road detection. Henry et al.~\cite{henry2018road} evaluated different FCNs for road segmentation in high-resolution synthetic aperture radar satellite images. Cheng et al.~\cite{cheng2017automatic} proposed an end-to-end cascaded convolutional network for road detection. The model consists of two networks. One aims at the road detection task and the other is cascaded to the former one, achieving full use of the feature maps, to obtain the good centerline extraction. In \cite{wei2020simultaneous}, the authors proposed a multistage framework to accurately extract the road surface and centerline simultaneously. This framework consists of two FCNs to boost the accuracy and connectivity of the image segmentation. 
\begin{figure*}
  \centering
  \includegraphics[width=0.78\textwidth]{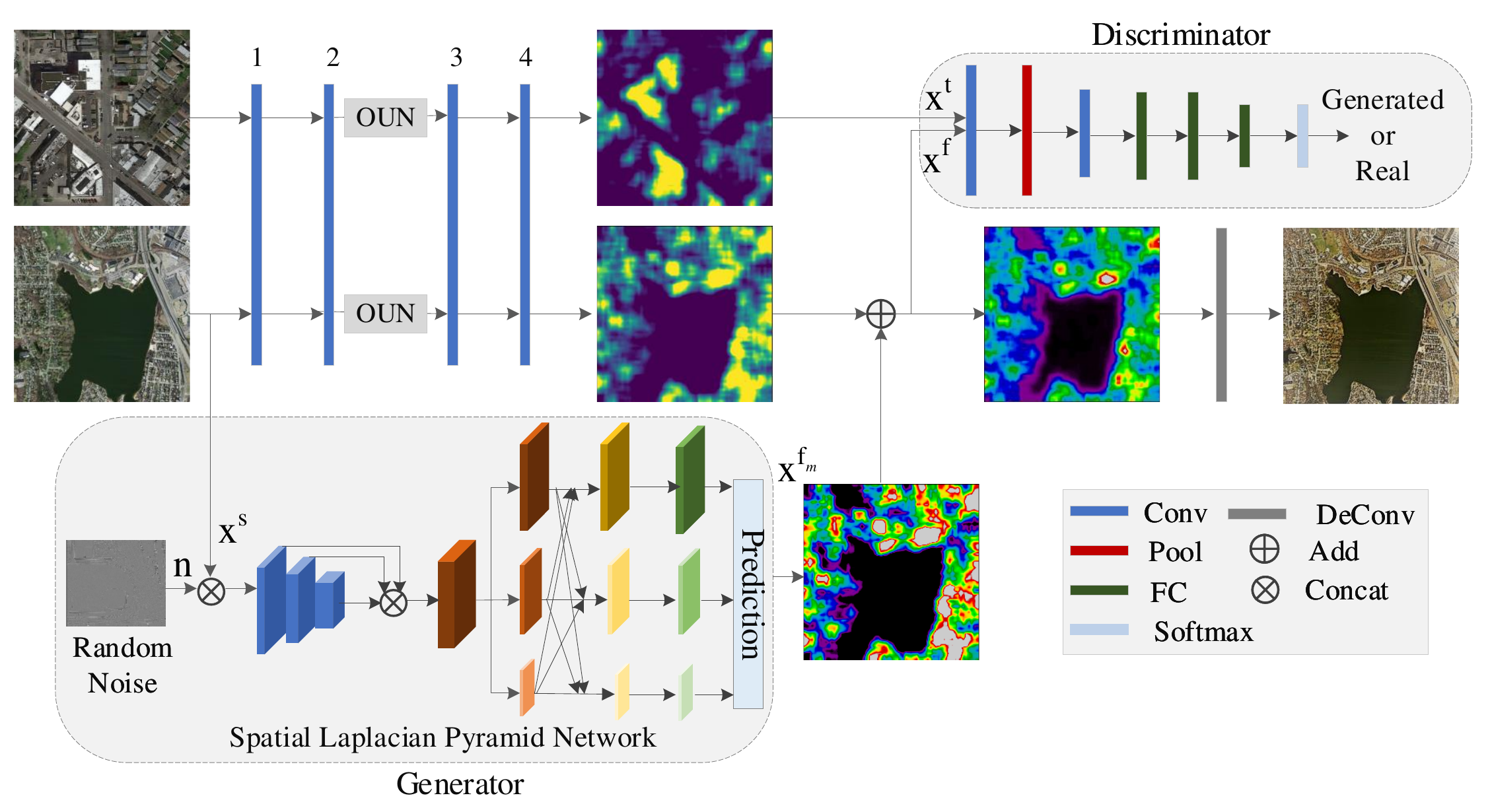}
\caption{\small The overall architecture of the proposed Pyramid Domain Adaption network.}
\label{fig:1}
\end{figure*}
\subsection{Domain Adaptation and Adversarial Networks}
DNN models are highly dependent on the training and testing data that have equal underlying distribution. Practically, it is common to have some diversity between the training and testing data. To rectify this difference and adjust the methods for better generalization, domain adaptation is recommended \cite{dong2019sig, teng2019classifier}. Hoffman et al.~\cite{hoffman2018cycada} proposed a cycle-consistent domain adaptation model that receives multiple forms of representations while enforcing local and global structural consistency through pixel cycle-consistency and semantic losses. The model uses a reconstruction loss to boost the cross-domain conversion to maintain structural information and a semantic loss to sustain semantic consistency. Zhu et.al.~\cite{zhu2017unpaired} proposed a learning method to translate an image from the source domain to the target domain by introducing cycle learning without paired examples. Wang et al.~\cite{wang2019weakly} proposed a weakly supervised adversarial domain adaptation method to improve the segmentation performance from the synthetic data, which consists of three networks. The detection model focuses on detecting objects and predicting a segmentation map; the pixel-level domain classifier attempts to distinguish the domains of image features; an object-level domain classifier discriminates the objects and predicts the objects classes. In \cite{li2019road}, the authors proposed a road segmentation model that combined the adversarial networks with multiscale context aggregation. In this approach, standard U-Net is used as the generator and FCNs are used as the discriminator. This model is computationally inefficient and has limited performance in handling complex images. In \cite{wang2018spectral}, the authors proposed a new linear space embedded clustering method, which uses adaptive neighbors to optimize the similarity matrix and clustering results simultaneously, also a linearity regularization is used to generate a linear embedded spectral from the data representation.

\subsection{Feature Pyramid Network}
For different tasks in high spatial resolution RS images, it has been proposed to use multiscale images for training. Feature pyramids are a module in recognition systems for detecting and segmenting of objects with different scales. Plenty of efforts have been made to improve performance accuracy. The authors in \cite{bellens2008improved} introduced a bottom-up model for RS image classification. To avoid the indirect classification and segmentation problems of these bottom-up methods, Sivic and Zisseman \cite{sivic2003video} proposed to recognize objects without classifying pixels or regions. Afterward, a spatial pyramid network is used for instance segmentation \cite{mao2019unsupervised, lu2017remote}. 
Recently, several feature pyramid architectures have been proposed that achieved encouraging results \cite{chen2019residual, zhang2019pan}, and still the current approaches have limitations as they just build the feature pyramids based on the original pyramidal architecture of the backbones with several scales. For instance, the authors in \cite{li2019nested} present a feature pyramid architecture for object detection in RS images with several dense blocks and pooling layers. Lin et al.~\cite{lin2017feature} presented a feature pyramid architecture for scale-wise object detection, the model has bottom-up and top-down paths to improve to the detection. 

Zhao et al.~\cite{zhao2019m2det} introduced a multi-level feature pyramid network to construct more accurate feature pyramids for identifying objects in different scales. Yu et al.~\cite{yu2018semantic} introduced a feature pyramid model based on pyramid pooling \cite{he2015spatial} and feature transformation. In \cite{li2019road} and \cite{yang2019cdnet}, the authors presented encoder-decoder pyramid networks based on sparse coding for cloud detection in an optimal and efficient way. The model extracts multiscale contextual information without the loss of resolution. Sun and Wu \cite{sun2019learning} proposed a model that used spatial pyramid attentive pooling for syntactic image generation that can be integrated into both generators and discriminators. Pang et al.~\cite{pang2019efficient} introduced a pyramid style model to produce featurized images in a single-stage detection framework. The multiscale features are then added into the prediction layers of the detector using an attention module. 

\section{PROPOSED METHOD}
\label{sec:3}
We intend to develop a road segmentation model in the source domain that has labeled datasets and generalizes the system to the target domain that has an unlabeled dataset. In this section, we first introduce the background of GANs and describe the design of the proposed ASPN model that has a domain adaptation structure. Then, we discuss the details of the proposed architecture.
\subsection{Background}
GANs generally contain a generator (G) that trace random noise, ($n$) to create samples and a discriminator, (D) that recognizes the fake samples from the real ones. The basic framework of the conditional GAN can be observed as a game between G and D, to figure out the equilibrium to the min-max problem,
\begin{eqnarray}
 \min_G \max_D E_{x~q(x)}[\log D(x)]+E_{n~p(n)}[\log(1-D(G(n)))]
\label{eq:1}
\end{eqnarray}
while $n\in R^{dn}$ is a hidden variable from the distributions of $N (0, 1)$ or $U[-1, 1]$. To generate a model based on the input images, DNNs are typically applied in both G and D. In the proposed architecture, the generator is conditioned based on the additional features maps $x^s$ of the synthetic images. While training, the generator 
$G(x^s, n; \theta_G)=x^s_{conv4}+\hat G(x^s, n; \theta_G)$  converts the synthetic feature maps $x^s$ and noise map $n$ to an adjusted feature map $x^{fm}$. It is noticeable that the residual presentation $\hat G(x^s, n; \theta_G)$ between the fourth convolution layer's feature maps of real and synthetic images, rather than straight-forward computation $x^{fm}$ is computed by the generator.

It is expected that $x^{fm}$ keeps the real semantics of the base feature map $x^s$. Hence, $x^f$ is fed to the discriminator $D(x,\theta_D)$ and to the classifier $T(x,\theta_T)$. Here, the task of discriminator $D(x,\theta_D)$ is to recognize the converted feature maps $x^{fm}$ that are created by the generator, from the real image's feature maps of the target domain $x^t$, at the last stage. 
The classifier division $D(x,\theta_T)$ assigns labels to each pixel in the input image, which is called the deconvolution layer \cite{pang2019efficient}. Fig.~\ref{fig:1} illustrates the overall structure of the proposed ASPN model. The objective is to optimize the following min-max function:
\begin{eqnarray}
\min_{\theta_G \theta_T}\max_{\theta_D}= \ell_d (G, D)+\partial \ell_t (G, T)
\label{eq:2}
\end{eqnarray}
while $\partial$ is a weight that operates the arrangement of the losses. $\ell_d$ denotes the domain loss:
\begin{eqnarray}
\ell_d (D, G)=  E_{x^t}[\log D(x^t, \theta_D)]   \nonumber\\
+ E_{x^s,n}[\log(1-D(G(x^s,n; \theta_G), \theta_D)) 
\label{eq:3}
\end{eqnarray}
In the deconvolution division, we introduce the function loss $\ell_f$ as a multinomial loss (cross-entropy loss):
$\setlength{\arraycolsep}{0.0em}$ 
\begin{eqnarray}
\ell_f (G,T)= E_{x^s, y^s, n}\Big[-\sum_{i=1}^{\vert P^s \vert} \sum_{k=1}^{S^c} 1^{y_i=k}\log(T(x_i^s, \theta_T)) \nonumber\\
 -\sum_{i=1}^{\vert P^s \vert} \sum_{k=1}^{S^c} 1^{y_i=k}\log (T(G(x_i^s,n,\theta_G),\theta_T)) \Big]  
\label{eq:4}
\end{eqnarray}
while $\sum_{i=1}^{\vert P^s \vert}$ and $\sum_{j=1}^ {S^c}$ determine the summary of $\vert P^s \vert$ pixels and $S^c$ semantic classes, $1^{y_i=k}$ is a hot encoding of the $i^{th}$ pixel. Multiscale pyramid networks with multiple transformations are used in G. Multi-layer perceptron is implemented in discriminator D. The min-max optimization is sought between two levels. Through the first level, the discriminator D and the pixel-wise classifier $T$ are frequently updated, while the conditional generator G and the feature extractor Convolution $1\sim4$ and a single OUN remain intact. Within the next level, T and D are stable while we update G, OUN, and Convolution $1\sim4$. It should be considered that $T$ is trained with both extracted and inferable  base feature maps. Training $T$ on the extracted feature maps particularly leads to similar efficiency, whilst demanding several rounds of initializations and multiple learning rates because of the stability issues of the GAN. In fact, if the model does not get trained on the source data, there is the possibility of the shift class assignments (e.g. class 1 change to 2), and the objective function remains optimized. In \cite{teng2019classifier}, the authors proposed to train classifier $T$ equally on the source and adapted images, which eliminate the class shifts and significantly stabilize training. In general, both G and D are cycle wise repeated while G accepts a random prior $z \in R^r$ to optimize the individual parameters, $\theta_g$ and $\theta_d$ are generated as follows,
\vspace{-3mm}
\begin{eqnarray}
\theta_d \leftarrow \theta_d + S \nabla\theta_d \frac{1}{m} \sum_{i=1}^m \log D(x_i)+\log(1- D(G(z_i)))  \nonumber\\
\theta_g \leftarrow \theta_g -  S \nabla\theta_g \frac{1}{m} \sum_{i=1}^m \log(1- D(G(z_i))) 
\label{eq:5}
\end{eqnarray}
in which $m$ is the minibatch size and $s$ is the step size. Indeed, G can be trained to take full advantage of $\log (D(G(z))$ rather than reducing $\log (1 - D(G(z))$ to deliver stronger gradients at the beginning of the training,
\begin{eqnarray}
\theta_g \leftarrow \theta_g -  S \nabla\theta_g \frac{1}{m} \sum_{i=1}^m \log (D(G(z_i)))
\label{eq:6}
\end{eqnarray}
Therefore, Eq.~\ref{eq:6} is applied in our model as its performance is more stable. By using the base network for pre-training, GAN has the ability to produce more distributed records. In this work, we propose OUN (the details are presented in the next section) that has encoding/decoding paths. Therefore, the pre-trained decoder $Dec$ can choose the appropriate pixels from G(z) for recovering the image $Dec(G(z))$. For the given input, D is trained to distinguish a synthetic sample $Dec(G(z))$ from a real one $x$. The proposed model is trained as follows,
\begin{equation}
\begin{aligned}
\theta_d \leftarrow \theta_d + S \nabla\theta_d \frac{1}{m} \sum_{i=1}^m \log D(x_i)+\log(1- D(x_i)) \\
\theta_{g, dec} \leftarrow \theta_{g, dec} -  S \nabla\theta_{g, dec} \frac{1}{m} \sum_{i=1}^m \log(D(x_i)) \\
in ~ which ~~~~~ x_{z_i} = Dec(G(z_i))
\label{eq:7}
\end{aligned}
\end{equation}
The aim of G is to generate samples that D is not able to distinguish them from the real ones. Therefore, G can learn to map different random priors $z$ into the same synthetic output, instead of creating diverse synthetic outputs.  This issue is denoted as "mode collapse" that happens when the GAN's optimization approach is solving the min-max problem. Minibatch averaging is used in our model, which is motivated by the philosophy of minibatch discrimination \cite{choi2017generating}. It provides D with the access to the minibatch of the real samples and the fake samples $(x_1, x_2, ..., and ~ G(z_1), G(z_2)$ respectively), whereas classifying samples. Given a sample to discriminate, minibatch computes the gap between the given sample and the other samples in the minibatch. Minibatch shows the average of the minibatch samples to D, hence the objective is modified as follows:
\begin{equation}
\begin{aligned}
\theta_d \leftarrow \theta_d + S \nabla\theta_d \frac{1}{m} \sum_{i=1}^m \log D(x_i, \overline x_z)+\log(1- D(x_{z_i}, \overline x_z)) \\
\theta_{g, dec} \leftarrow \theta_{g, dec} -  S \nabla\theta_{g, dec} \frac{1}{m} \sum_{i=1}^m \log(D(x_{z_i}, \overline x_z)) \\
in ~  ~ which ~~ \overline x = \frac{1}{m}Dec(G(z_i)), \overline x_z = \frac{1}{m}\sum_{i=1}^m x_{z_i}
\label{eq:8}
\end{aligned}
\end{equation}
while $m$ represents the size of the minibatch. Consequently, the average of the minibatch is concatenated on the provided samples and fed to D.
\begin{figure*}
  \centering
  \includegraphics[width=0.78\textwidth]{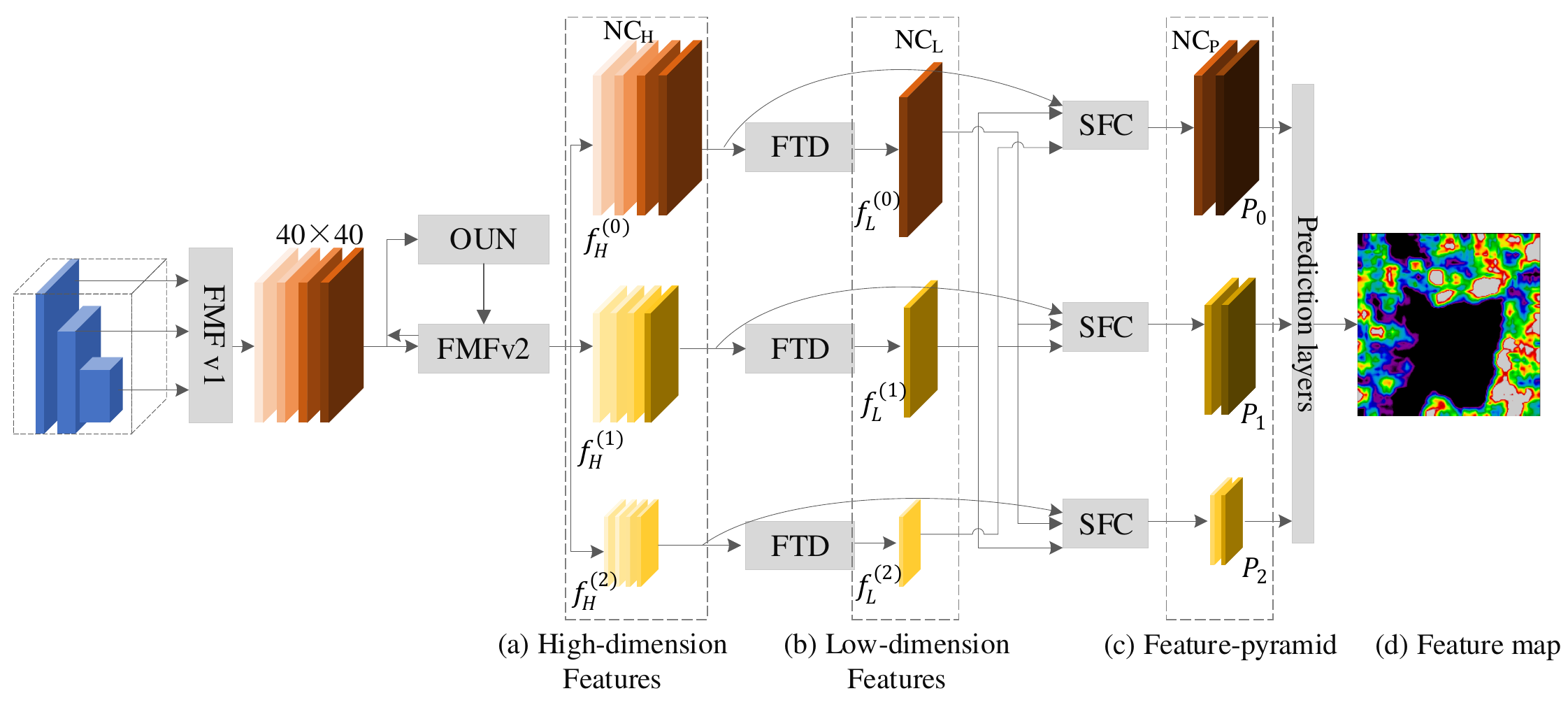}
\caption{\small The architecture of the proposed feature pyramid network.}
\label{fig:2}
\end{figure*}
\subsection{Generator Architecture}
\begin{figure*}
  \centering
  \includegraphics[width=0.75\textwidth]{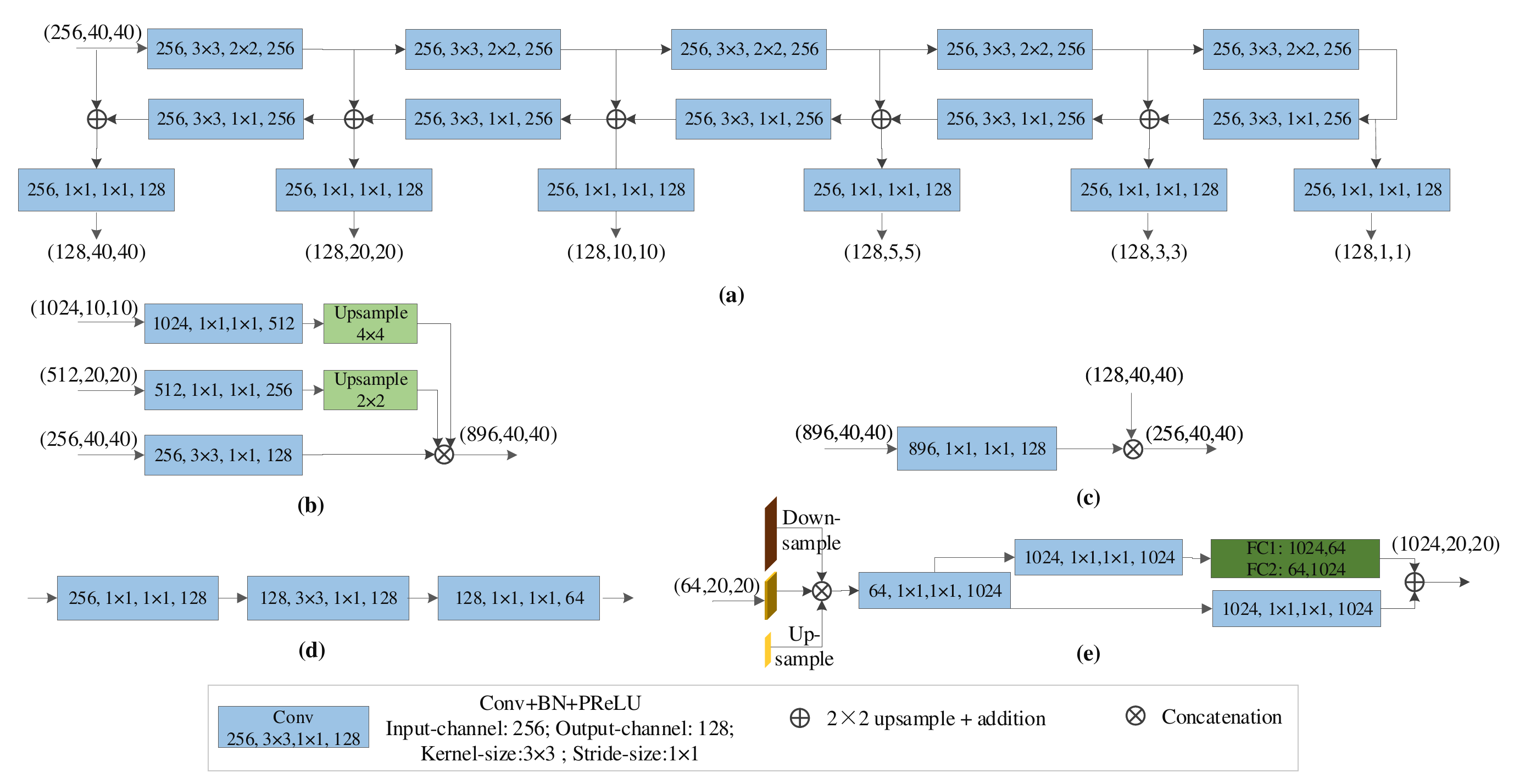}
\caption{\small The detailed structure of several modules. (a) OUN, (b) FMFv1, (c) FMFv2, (d) FTD, (e) SFC. The numbers inside the blocks denote input channels, Conv kernel size, stride size, output channels.}
\label{fig:3}
\end{figure*}

In particular, the aim of the generator is to produce images similar to real ones for synthetic images to minimize the domain gap. To achieve this goal, we propose a feature pyramid network with multiple connections to augment the representations of the synthetic images, as well as several transformations that are helpful for detecting multiscale objects and generating accurate feature maps. The feature maps in the feature pyramid (FP) that are based on the bottom-up convolution network cannot fulfill the demands of the upper-level feature maps that are achieved by the deep transformation functions; in contrast, the lower-level feature maps acquired by the shallow transformation units also cannot deliver across the second property that limits the detection performance on small objects. Moreover, each generated feature map just represents the output of its corresponding scale, so relative information of the other scales cannot be successfully compounded. One of the options to prevail over these limitations is to applying the transformation functions with an appropriate depth to preserve the sufficient spatial information as well as the high-level semantic information.
The architecture of the proposed FP network is shown in Fig.~\ref{fig:2}. In the propose model, the backbone and the Multi-stage Feature Pyramid Network are used to extract features, followed by the non-maximum suppression (NMS) operation to generate feature maps from the input images. The proposed FP network contains four divisions. Feature Map Fusion (FMF), Optimized U-shape Network (OUN), Feature Transportation Division (FTD) and Scale-wise Feature Concatenation (SFC). FMFv1 enhances the semantic information of the basic features by combining the feature maps of the backbone. The OUN largely produces a collection of multiscale features, and then the FMFv2 is applied to fusing the features produced by OUN with the basic features to extract multi-stage multiscale features. The concatenated feature maps again are fed to the next OUN (there are six groups of OUN+FMFv2). The first OUN is just learned from $X_{ini}$. The final multi-stage multiscale output features are computed as follows:
\begin{equation}
\begin{aligned}
\big[ X_1^l, X_2^l, ..., X_i^l\big] = 
\begin{cases}
T_l (X_{ini}) & \text{if}~~{ l = 1}\\
T_l F (X_{ini}, X_i^{l-1}) & \text{if} ~~ {l = 2, ..., \' L}
\end{cases}
\label{eq:9}
\end{aligned}
\end{equation}
while $X_{ini}$ represents the initial features that are extracted from the base network, $X_i^l$ signifies the features with the $i^{th}$ scale in the $l^{th}$ OUN, $L$ represents the total number of OUNs, $T_l$ shows the performance of $l^{th}$ OUN, and $F$ represents the proceeding of FFMv1. Furthermore, SFC gathers the multistage feature pyramid via a scale-wise feature interpolation function and an adjusted attention approach. The details of each division is shown in Fig.~\ref{fig:3} and illustrated as follows.

{\bf Base Network:} Similar to \cite{chen2019residual}, ResNet-101 and VGGNet-16 are adopted as the base network. In the proposed FP network, the last FC layers of ResNet-101 and VGGNet-16 are switched with a convolution layer for sub-sampling their parameters. In the base network, the output of the $l^{th}$ layer is signified as $b_l$ and the outputs of the backbone network are presented as $B_{net}= \{b_1, b_2,..., b_L\}$, accordingly the prediction feature map sets can be represented as follows,
\begin{equation}
B_{pred}=\{b_P, b_{P+1}, ..., b_L\}
\label{eq:10}
\end{equation}
while $P\gg l^3$ , $b_L$ are deep feature maps. When $P< l< L$, $b_l$ goes to the shallow feature maps and extracts low-level features. The high-resolution maps with partial semantic information may not well lead to object detection and segmentation. In our observation, reusing deep and shallow semantic information is the key bottleneck in increasing the performance of the model. To enhance the shallower layers' semantic information, we can add the features from the deeper layers. For example,
\begin{equation}
\begin{aligned}
\acute b_L &= b_L &&\\
\acute b_{L-1} &= w_{L-1}.b_{L-1}+\alpha_{L-1}.b_L,&&  \\
\acute b_{L-2} &=w_{L-2}.b_{L-2}+\alpha_{L-2}.\acute b_{L-1},&& \\
&= w_{L-2}.b_{L-2}+\alpha_{L-2} w_{L-1}.b_{L-1}+\alpha_{L-2} \alpha_{L-1}.b_L &&
\label{eq:11}
\end{aligned}
\end{equation}
where $w$ and $\alpha$ are weights. Without considering the generalization loss,
\begin{equation}
\acute b_l = \sum_{i=p}^L W_l. b_l
\label{eq:12}
\end{equation}
where $W_l$ denotes the generated weights for the output of the $l^{th}$ layer and the final features are expressed as:
\begin{equation}
\acute B_{pred} = \{\acute b_P, \acute b_{P+1}, ..., \acute b_L\}
\label{eq:13}
\end{equation}
From Eq.~\ref{eq:12}, it can be observed that the final features $(\acute b_l)$ are corresponding to the merging $(b_l, b_{l+1},…,b_L)$. One of the methods to enhance the shallow layers's information is the linear combination with the deeper feature hierarchy. 

{\bf FMF:} The task of this division is to fuse the features that are extracted from different levels, and have a rule in constructing the last multi-stage pyramid feature map. Firstly, $1\times 1$ convolution layers are used to compact all the channels of the input features and secondly, to combine these feature maps, concatenation is applied. In particular, FMFv1 receives three different scales' feature maps from the backbone network as the input and it has two different scale upsampling functions to rescale the deep features to the equal scale before concatenating the features. On the other hand, FMFv2 receives two same scale feature maps as the input. One is the base feature and the other one is the largest output feature map of the earlier OUN and we generate the fused feature for the next OUN. The detailed structures of these two divisions are presented in Fig.~\ref{fig:3}(b) and (c), respectively.

{\bf OUN:} In contrast to other FP networks \cite{zhao2019m2det, chen2019residual}, in the proposed FP network, OUN is used, and the details are presented in Fig.~\ref{fig:3}(a). The down-sampling path has five sequences of $3\times 3$ convolution layers with stride 2. The up-sampling path receives the outcomes of different layers as its reference set, however, the other approaches just receive the result of the final layer of each stage in the residual network backbone \cite{pang2019efficient}. Furthermore, to increase the learning performance and preserve the smoothness for the features, $1\times1$ convolution layers are added after each up-scaling and addition process in the up-sampling path. In each OUN (six are used), all the outputs in the up-sampling path configure the multiscale features at the present stage. Overall, the stacked of OUNs build the multi-stage multiscale features, from the shallow to the deep level features. 

{\bf FTD:} In FTD, for computational efficiency, three convolution layers are implemented with the size of $1\times1, 3\times3$ and again $1\times1$ to decrease the channel number. Moreover, for input normalization and activation, the batch normalization (BN) and the parametric rectified linear unit (PReLU) are used. $1\times1$ convolution in the FTD reduces the special feature maps by half channels $NC_H= D$, while NC is the number of the output channels of the FTD. 

{\bf Feature Pyramid Pooling}. In classification and segmentation tasks, pooling layers are widely used \footnote{\url{http://deepglobe.org/, 2018.}}, and these layers not only spatially decrease the size of the feature maps, but also combine the contextual information of the sub-regions. He et al.~\cite{he2015spatial}, introduced a model that use different sub-regions pooling sizes to generate feature pyramids (FPs) for object detection and segmentation. If the base network generates the $W\times H$ size feature map with $D$ channels, firstly to each high dimensional FP, pooling is applied, $F_H=\{f_H^{(0)},f_H^{(1)},…,f_H^{(N-1)}\}$, while $f_H^{(n)}$, with the feature map spatial size of $\frac{W}{2^n}\times \frac{H}{2^n}$, represents the $n^{th}$ stage of $F_H$, and $N$ signifies the level of the pyramids. Hence, the down-sampling of the feature maps is used to reduce the spatial size by half. Later, the FTD is used to decrease the channel numbers. Meanwhile, the output of the FTD with low-dimensional FP pooling is represented as $F_L = \{F_L^{(0)}, F_L^{(1)},…,F_L^{(N-1)}\}$ while the reduced channel number is $C_L = D/(N-1)$. 

{\bf SFC:} The rule of this layer is to combine the multiscale features that are stepwise produced by OUNs to generate the multiscale feature pyramid maps. The details are presented in Fig.~\ref{fig:3}(e). Firstly, the SFC concatenates the same scale features together with the channel dimension. As the SFC units reuse the feature maps in $F_L$, efficiently, we can extract the different scales' context information to derive $P_n$ with $NC_P$ channels. In SFC, the feature maps from $F_H$ are aggregated with feature maps of $F_L$ by the help of skip connection. Hence, the numbers of the output channels reach to $2D/N$, therefore, the aggregated feature maps have $2D$ channels. The concatenated feature pyramids can be denoted as $X=[X_1, X_2, ..., X_i]$, while $X_i = Concat (X_i^1, X_i^2, ..., X_i^L)\in R^{(W_i\times H_i\times NC)}$ states the features of the $i^{th}$ largest size. After applying the aggregation to different scale feature pyramids, the output of this division contains features from several layers depths. However, just applying simple aggregations is not sufficient. Therefore, the channel-wise attention module is used to enforce the features to focus on the most representative channels. Global average pooling \cite{hu2018squeeze} is used to create channel-wise statistics $GP\in R^{NC}$ at the combining section. To precisely capture the channel-wise dependencies, two fully connected layers are used as follows:
\begin{equation}
A = F_{ex}(GP, W) = \rho (W_2^\sigma (W_1^{GP}))
\label{eq:14}
\end{equation}
while $\rho$ denotes the PReLU function, $\sigma$ represents the softmax function, $W_1\in R^{\frac{NC}{r} \times NC}$, $W_2\in R^{NC\times \frac{NC}{r}}$, $r$ denotes the decreasing ratio (in this work $r = 8$). The last output is computed by reweighting the input $X$ with activation $A$, while $\tilde X_i = [\tilde X_i^1, \tilde X_i^2,..., \tilde X_i^c]$ and the rescaling process either helps the features to become more accurate or weakened.
\begin{equation}
\tilde X_i^{NC} = F_{scale} (X_i^{NC}, A_{NC}) = A_{NC}.X_i^{NC}
\label{eq:15}
\end{equation}

Prior to the whole network training, we pre-train the backbone on the datasets. The input size to the proposed FP network is $320\times 320$ and the network contains $6$ OUNs, where each OUN has $5$ convolution layers in the down-sampling path and $5$ in the up-sampling path, and therefore the output features have $6$ different scales. To decrease the number of the parameters, we only assign $256$ channels to each scale of their OUN features; hence, the network can simply train on GPUs. Two convolution layers are added to each of the $6$ pyramidal features to achieve location regression and perform classification. At every pixel of the pyramidal features, six anchors with three ratios are set and a probability score (threshold) of $0.05$ to cut out anchors that have low scores. Reducing the threshold to $0.01$ can improve the performance; however, it also increases the processing time.
\subsection{Discriminator Architecture}
As Fig.~\ref{fig:1} shows, the discriminator is trained to distinguish the created feature for the source domain images and the real ones from the target domain images. It receives the vectorized feature maps as input and pass them to the three FC layers followed by a classification layer with the softmax function. The output dimensions of the three FC layers are $4096$, $4096$ and $1024$ respectively. By differentiating between the fake and real image representations, an adversarial loss is proposed to help the generator to create the representation for the synthetic images that are quite similar to the genuine ones.
\subsection{Real Images Testing}
In the testing part, we used our previous work reported in \cite{shamsolmoali2019novel}. A real image goes over the feature extractor, Convolution and OUN layers followed by the pixel-wise classifier for road segmentation \cite{shamsolmoali2019novel}. The generator and discriminator will not participate and the proposed architecture should almost have similar interpretation complexity. The model achieves $4.48$ fps with one NVIDIA GTX GeForce $1080$ Ti GPU. To discuss the computation cost of ASPN, first, we consider a two scale pyramid network \cite{pang2019efficient}. Here $I (j\times j)$ denotes an image, $l = d(I)$ represents the coarsened image, and $h = I - u(d(I))$ calculates the high pass. To streamline the computations, a different $u$ operator is used to produce the images. We have taken $d(I)$ as the mean of each disjoint block of $2\times 2$ pixels, and $u$ as the factor that eliminates the mean from all the $2\times 2$ blocks. Subsequently, $u$ has the rank of $3d^2/4$, here, we assign $h$ as an orthonormal basis to the range of $u$, therefore the linear mapping between $I$ and $(l, h)$ is unitary. For creating a probability density $p$ on $R^{d^2}$, we follow, 
\begin{equation}
p(I) = q_0 (l, h) q_1 (l) = q_0 (d(I), h(I)) q_1 (d(I));
\label{eq:16}
\end{equation}
if $q_i\geq 0$, $\int_{a}^{b} q_1 (l)dl = 1$, and for every constant $l$, $\int_{a}^{b} q_0 (l, h)dh = 1$ . To verify whether or not $p$ has a unit integral:
\begin{equation}
\begin{aligned}
\int pdI &= \int q_0 (d(I), h(I))~q_1(d(I))~dI&& \\
& = \int \int q_0 (l, h)~q_1 (l)~dl~dh =1 &&
\label{eq:17}
\end{aligned}
\end{equation}
A set of training samples $(l_1, ..., l_{N0})$ are taken for $q_1$ , and the density function is accordingly constructed as follows:
\begin{equation}
q_1 (l) \sim \sum_{i=1}^{N_1} e^{\Vert l - l_i \Vert^2 / \sigma_1}    
\label{eq:18}                              
\end{equation}
To define $q_0 (I) = q_0 (l, h)\sim \sum_{i=1}^{N_0} e^{\Vert l - l_i \Vert^2 /\sigma_1}$ , we set $l = d (I)$.  A similar method is used for each of the finer scales while the pyramid network has more levels. It should be considered that generally the  low pass filter at each scale is used, and additionally we measure the true high pass versus the generated high pass samples of the model. Therefore, using a pyramid that has $M$ levels, the last log likelihood is computed as follows:
\begin{equation}
\log(q_M (l_M)) + \sum_{M=0}^{M-1} \log(q_M(l_M, h_M)). 
\label{eq:19}
\end{equation}
The training pseudocode of ASPN is shown as follows.

\begin{algorithm}
\caption{ASPN Training Pseudocode}
\small{
{\bf for} number of training iterations  {\bf do} \\
~ ~ {\bf for} k steps  {\bf do}   \\
~ ~ ~ $\bullet$  Draw a minibatch of samples $\{x_1, ..., x_m\}$ from \\ ~ ~ ~ ~ generated data distribution. \\
~ ~ ~ $\bullet$  Draw a minibatch of noise samples $\{G_{z_1}, ..., G_{z_m}\}$ \\ ~ ~ ~ ~ from noisy prior $(n)$. \\
~ ~ ~ $\bullet$  update the discriminator: \\
~ ~ ~ ~ ~ $s \nabla \theta_d \frac{1}{m} \sum_{i=1}^m \log D(x_i, \bar x) + \log (1 - D(x_{z_i}, \bar x_z))$ \\
~ ~ {\bf end for} \\
~ ~ ~ $\bullet$  $Dec$ is the pre-trained decoder which chooses the \\ ~ ~ ~ ~ appropriate pixels from $G(z)$ for image recovery\\ ~ ~ ~ ~ ~ $(x_{z_i} = Dec (G(z_i)))$.     \\        
~ ~ ~ $\bullet$   update the generator:       \\
~ ~ ~ ~ ~ ~ $s \nabla \theta_{g, dec} \frac{1}{m} \sum_{i=1}^m \log D(x_{z_i}, \bar x_z))$ \\    
~ ~ ~ $\bullet$  Update the feature extractor, source domain classifier \\ ~ ~ ~ ~ and target domain classifier: \\
~ ~  ~ ~  ~  $l_f (G, T) = E_{x^s, y^s, n} \Big[-\sum_{i=1}^{\vert p^s \vert} \sum_{k=1}^{s^c} 1^{y_i = k}$ \\  
~ ~ ~ ~  ~ $\log(T(x_i^s, \theta_T))$ $-\sum_{i=1}^{\vert P^s \vert} \sum_{k=1}^{s^c} 1^{y_i = k}$ \\ ~ ~ ~ ~ ~  $\log(T(G(x_i^s, n, \theta_g), \theta_T)) \Big]$ \\
{\bf end for} \\
We use $\theta_g \leftarrow \theta_g - s \nabla \theta_g \frac{1}{m} \sum_{i=1}^m \log(D(G(z_i)))$ to deliver stronger gradients at the beginning of the training.
}
\label{algo:1}
\end{algorithm}

\section{EXPERIMENTAL DETAILS AND ANALYSIS}
In the following section, we explain the details of the used datasets, experimental results, and compare the performance of the ASPN model with that of the other models.
\vspace{-4mm}
\subsection{Datasets and Experimental Setup}
To accurately evaluate the performance of the proposed model, three road detection datasets are used in our experiments. These datasets have different distributions and densities of objects captured from different distances. To improve the diversity of the training samples, we have the following alterations for data augmentation.
\begin{enumerate}
\item Random vertically and horizontally flipping. 
\item Random conversion by $[-8, 8]$ pixels. 
\item Random scaling in the range $[1, 1.5]$.
\end{enumerate}  
\vspace{1mm}

\noindent {\bf DeepGlobe road extraction dataset \footnote{\url{http://deepglobe.org/, 2018.}}:} It is a 2-tiles dataset from India, Thailand, and Indonesia. The road extraction dataset has complex road environments. The data set holds $6226$ training images and $2344$ testing and validation images. Each image has a size of $1024\times 1024$, and the ground resolution of the image pixels is $0.5m/pixel$. We used all the training samples to pre-train the backbone and $1169$ samples for training the network and ignored repetitive images.\par
\noindent {\bf Massachusetts road dataset \footnote{\url{https://www.cs.toronto.edu/~vmnih/data/, 2013}}:} The dataset consists of $1171$ images of the state of Massachusetts. The size of each image is $1500\times1500$ pixels with a spatial resolution of $1m$ per pixel, composed of red, green and blue channels. This dataset was collected from aerial images. The ground truth of the images consists of two classes, roads and non-roads. \par
\noindent {\bf EPFL Road Segmentation dataset \footnote{\url{https://www.kaggle.com/c/epfml17-segmentation/leaderboard, 2017.}}:} The dataset consists of $150$ aerial images. The size of each image is $400\times400$ pixels with a spatial resolution of $1$m per pixel. \par
The Massachusetts validation and test sets are used for the proposed model validation and testing. To train the model, random mini-batches are selected from the images (and their labels) of the source domain dataset and the real images (without labels) from the target domain dataset. In addition, $20$ images from the testing set of the EPFL road dataset are used to test ASPN. The proposed model is evaluated based on Fr\'echet Inception Distance (FID) \cite{heusel2017gans}. The samples from $X$ and $Y$ are inserted into a feature space by using a particular Inception network. These feature distributions are formed as multidimensional Gaussians parameterized by their individual covariance and mean. Then the FID is measured by,
\begin{equation}
FID=\Vert \mu_x - \mu_y \Vert^2 + Tr\big(\sum_x + \sum_y -2(\sum_x \sum_y)^{\frac{1}{2}}\big)
\label{eq:20}                              
\end{equation}
while $(\mu_x, \sum_x)$ and $(\mu_y, \sum_y)$ represent the mean and covariance distribution of the real and generated images respectively. We adopted the FID score to evaluate the performance of the generative model on the unlabeled data. Additionally, the code released with the Cityscapes dataset \cite{hong2018conditional} is also used to evaluate the results. It computes the intersection-over-union $=\frac{TP}{(TP+FP+FN)}$, while TP, FP, and FN are the numbers of true positive, false positive, and false negative pixels, respectively that are calculated over the test set. Additionally, we used a sematic instant-level Intersection-over-union metric (iIoU), which represents how well each instance in the scene is signified in the labeling.\par
To fairly evaluate the performance of ASPN, several evaluation metrics have been chosen, such as the Variation of Probabilistic Rand Index (PRI) \cite{unnikrishnan2007toward}, Information metric (VOI) \cite{meila2005comparing}, Global Consistency Error (GCE) \cite{martin2001database}, and Boundary Displacement Error (BDE) \cite{freixenet2002yet}. Based on these parameters, if a segmentation result is good, whenever the comparison with ground truth yields a high value for PRI and small values for the other three metrics. PRI can be defined in a simple form. Let $S_{ground}$ and $S_{test}$ be two clusters of the same image, and $n_{ij}$ is the sum of points in the $i^{th}$ cluster of $S_{ground}$ and the $j^{th}$ cluster of $S_{test}$. $N$ signifies the total number of the pixels of the image. The value of PRI, which measures the similarity of two clusters, ranges between $0$ and $1$ and can be formulated as follows:
\begin{equation}
\begin{aligned}
PRI(S_{ground}, S_{test}) =\Big\{\binom{n}{2} - 1/2 \big\{\sum_i (\sum_j n_{ij}^2) \\
+ \sum_j (\sum_i n_{ij}^2) - \sum \sum n_{ij}^2 \big\} \Big\} / \binom{n}{2}
\label{eq:21}
\end{aligned}
\end{equation} 
VOI, computes the loss/gain between the images. $H$ and $I$ respectively denote the entropies and the mutual information between the two clusters, and defined as:
\begin{equation}
VOI(S_{ground}, S_{test}) = H_{ground}+ H_{test} -2I (S_{ground}, S_{test})
\label{eq:22}
\end{equation}
GCE, defines the consistency of segmentation. $R(S_{ground}, p_i ) \Delta R(S_{test}, p_i)$ represents the symmetric difference between $R(S_{ground}, p_i)$ and $R(S_{test}, p_i)$. The non-symmetric local consistency error is calculated as:
\begin{equation}
E(S_{ground}, S_{test}, P_i) = \frac{R(S_{ground}, p_i)\Delta R(S_{test}, p_i)}{R(S_{ground}, p_i)}
\label{eq:23}
\end{equation} 
GCE is computed by symmetrization and averaging
\begin{equation}
\begin{aligned}
GCE(S_{ground}, S_{test}) = \frac{1}{N}min \Big\{\sum_i E(S_{ground}, S_{test}, P_i), \\
\sum_i E(S_{test}, S_{ground}, P_i)\Big\}       
\label{eq:24}
\end{aligned}
\end{equation} 
BDE measures the average displacement error of boundary pixels on the segmentation outputs. Indeed, it states the error of one boundary pixel as the distance between the pixel and its nearby boundary pixel in the image.
\begin{equation}
d(p_i, B_2) = \Vert p_i - p\Vert_{p\in B_2}^{min}
\label{eq:25}
\end{equation} 
The distance of a boundary point to the boundary set $B_2$ represented by $p_i\in B_1$ and $N_1, N_2$ denote the total number of the points in the boundar$y$ sets $B_1$ and $B_2$, accordingly BDE is defined as:
\begin{equation}
BDE(B_1, B_2 ) = \frac{\sum_i^{N_1} d(p_i, B_2) / N_1 + \sum_i^{N_2} d(p_i, B_1) / N_2}{2}
\label{eq:26}
\end{equation} 
\begin{table*}
\centering
  \caption{\small{PERFORMANCE COMPARISONS {\footnotesize{IN}} TERMS {\footnotesize{OF}} MEAN IOU {\footnotesize{AND}} iIOU PERCENTAGE, FID, NUMBER {\footnotesize{OF}} PARAMETERS (Param) {\footnotesize{AND}} COMPUTATION SPEED (FLOPs) {\footnotesize{ON}} MASSACHUSETTS ROAD TEST DATASET.}}
  \label{tab:1}
  \begin{tabular}{llclccccc}     \hline
\\ [-0.5em]
    Methods & Backbone & Input size & Strategy & IoU & iIoU & FID & Param & FLOPs \\     \hline
\\ [-0.5em]
Xu and Zhao \cite{xu2018satellite} & FCN & $512\times512$ & Residual learning & 52.34 & 40.97 & 31.2 & 13.37M & 84.28B \\
Huang et al.~\cite{huang2013unified} & --- & $800\times800$ & Unified fusion &	55.62	& 43.71 & 28.6 & 13.51M & 84.75B \\
Zhu et al.~\cite{zhu2017unpaired} & FCN & $256\times256$ & discriminator update base & 71.77 &	60.16& 26.63 & 15.32M & 86.98B \\
Shrivastav et al.~\cite{shrivastava2017learning} & ResNet & $55\times35$ & discriminator update base & 73.53 & 62.23 & 20.42 & 14.82M & 86.44B \\
Hoffman et al.~\cite{hoffman2018cycada} & FCN & $600\times600$ & Cycle-Consistent adaptation & 75.15 &	 63.03 & 19.83 & 15.49M & 87.67B \\
Sun and Wu \cite{sun2019learning} & FCN &	$256\times256$ & Pyramid attentive pooling &	75.25	& 63.13 & 19.76 & 14.62M & 86.41B  \\
Hong et al.~\cite{hong2018conditional} & VGG-19 & $480\times960$ & Domain adaptation	&75.48 & 63.28 & 19.52 & 14.78M & 86.73B  \\
Liu et al.~\cite{liu2018roadnet} & FCN & $512\times512$ & Deep U-shape & 78.67 & 63.59 & 18.53 & 14.97M & 86.85B \\
Li et al.~\cite{li2019nested} & --- & $800\times600$ & Two-Stream Pyramid & 78.72 & 63.65 & 18.52 & 14.96M & 86.87B  \\
ASPN (Ours) & FCN& $512\times512$ & Domain adaptation & 74.81 & 62.31 & 20.27 & 14.32M & 86.23B  \\
ASPN (Ours) & FCN	& $320\times320$ & Domain adaptation	& 74.68 & 60.95 &	20.36 & 14.06M & 85.89B  \\
ASPN + 4 OUNs &ResNet	& $320\times320$	&Multiscale Domain adaptation	&77.79	&64.56	&18.79	&14.43M	&86.52B  \\
ASPN + 6 OUNs &ResNet	& $320\times320$	&Multiscale Domain adaptation	&77.96	&63.81	&18.75	&14.54M	&86.68B  \\
ASPN + 8 OUNs &ResNet	& $320\times320$	&Multiscale Domain adaptation	&77.92	&63.80	&18.78	&14.62M	&86.71B  \\
ASPN + 4 OUNs &VGG-16	& $320\times320$	&Multiscale Domain adaptation	&77.85	&62.73	&18.69	&14.34M	&86.26B  \\
ASPN + 6 OUNs &VGG-16	& $320\times320$	&Multiscale Domain adaptation	&77.99	&62.94	&18.66	&14.51M	&86.47B  \\
ASPN + 8 OUNs &VGG-16	& $320\times320$	&Multiscale Domain adaptation	&77.97	&62.93	&18.68	&14.67M	&86.53B  \\
ASPN + 6 OUNs &ResNet	& $512\times512$	&Multiscale Domain adaptation	&78.34	&63.37	&18.60	&15.03M	&87.22B  \\
ASPN + 8 OUNs &ResNet	& $512\times512$	&Multiscale Domain adaptation	&78.31	&63.34	&18.65	&15.11M	&87.3B  \\
ASPN + 6 OUNs &VGG-16	& $512\times512$	&Multiscale Domain adaptation	&78.86	&63.74	&18.43	&14.89M	&86.78B  \\
ASPN + 8 OUNs &VGG-16	& $512\times512$	&Multiscale Domain adaptation	&78.85	&63.72          &18.51	&14.96M	&86.84B  \\
        \hline
\end{tabular}
\end{table*}
\begin{figure*}
\vspace{-1cm}
  \centering
  \includegraphics[width=0.75\textwidth]{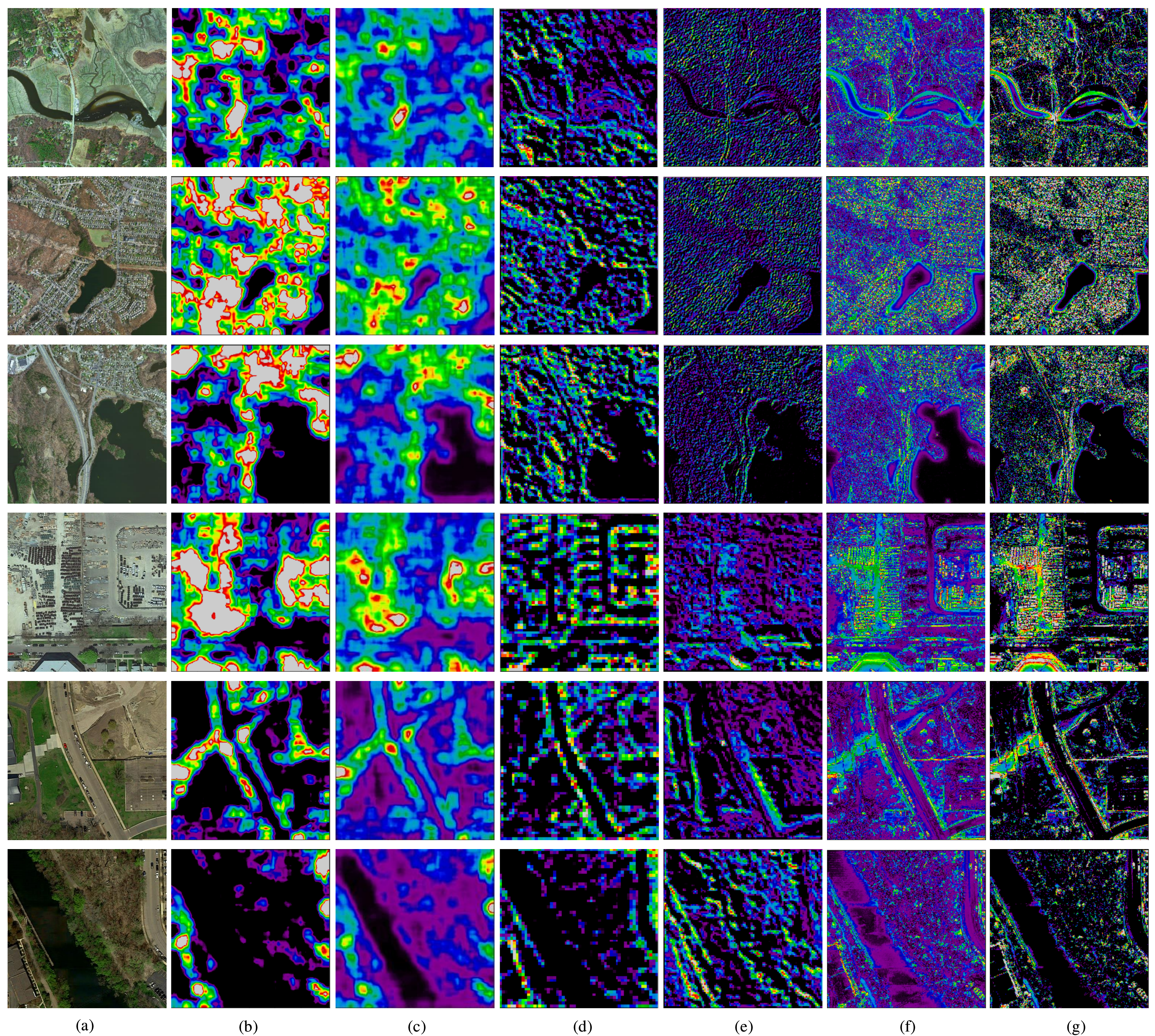}
\caption{\small Samples from synthesized images and visualization of attention map in different modules. (a) Synthesized image. (b) Attention map for $C_4$ + feature pyramid. (c) Attention map for $C_1$ + feature pyramid. (d) Attention map for $C_4$. (e) Attention map for $C_3$. (f) Attention map for $C_2$. (g) Attention map for $C_1$.}
\label{fig:4}
\end{figure*}
\vspace{-4mm}
\subsection{Implementation}
In our conditional GAN experiment, we use ResNet-101 \cite{he2016deep} as our backbone network. We initialize it with VGGNet-16 \cite{simonyan2014very}, and the Adam optimizer \cite{kingma2014adam} is used for training. The mini-batches contain $5$ images from the source domain and $5$ from the target one. ASPN is implemented by using Keras $2.1.2$, the deep learning open-source library and TensorFlow $1.3.0$ GPU as the backend deep learning engine. Python $3.6$ is used for all the implementations. All the implementations of the network are conducted on a workstation equipped with an Intel $i7-6850K$ CPU, a $64$ GB Ram and four NVIDIA GTX Geforce $1080$ Ti GPU and the operating system is Ubuntu $16.04$. All the images are resized to $320\times320$, thus, the feature map of the first convolution layer output is $[64, 320, 320]$. Therefore, $n$ is a $320\times320$ matrix, which is sampled based on a regular distribution $n_{ij}$$\sim$$u(-1,1)$. The first convolution layer feature map is concatenated to $n$ as an additional channel, and fed to the proposed pyramid network. To train the proposed pyramid network with the ResNet-101 backbone, the whole training time costs $5$ days and with the VGG-16 backbone, the entire training time costs $2.5$ days. 
\begin{table*}
\centering
  \caption{\small{EVALUATION {\footnotesize{OF}} CONDITIONAL GENERATOR}}
  \label{tab:2}
\small{
  \begin{tabular}{lcccccc}     \hline
\\ [-0.5em]
    Datasets & & Hoffman et al.~\cite{hoffman2018cycada} & Hong et al.~\cite{hong2018conditional}& Li et al.~\cite{li2019road} & ASPN Wo G &ASPN W G   \\     \hline
\\ [-0.8em]
\multirow{2}{*}{EPFL road dataset} & \multicolumn{1}{l}{IoU} & \multicolumn{1}{l}{62.3} & \multicolumn{1}{l}{68.72} & \multicolumn{1}{l}{71.22} & \multicolumn{1}{l}{48.53} & \multicolumn{1}{l}{81.68} \\[-0.5ex]
                                                           & \multicolumn{1}{l}{iIoU} & \multicolumn{1}{l}{50.9} & \multicolumn{1}{l}{54.26} & \multicolumn{1}{l}{56.37} & \multicolumn{1}{l}{39.67} & \multicolumn{1}{l}{67.43}  \\ \cline{1-7}
\\ [-0.8em]
\multirow{2}{*}{Massachusetts road dataset} & \multicolumn{1}{l}{IoU} & \multicolumn{1}{l}{58.63} & \multicolumn{1}{l}{76.18} & \multicolumn{1}{l}{77.15} & \multicolumn{1}{l}{44.25} & \multicolumn{1}{l}{78.86} \\[-0.5ex]
                                                           & \multicolumn{1}{l}{iIoU} & \multicolumn{1}{l}{45.37} & \multicolumn{1}{l}{62.48} & \multicolumn{1}{l}{62.86} & \multicolumn{1}{l}{36.78} & \multicolumn{1}{l}{63.74}  \\ \cline{1-7}  
\\ [-0.8em]
\multirow{2}{*}{DeepGlobe road dataset} & \multicolumn{1}{l}{IoU} & \multicolumn{1}{l}{55.34} & \multicolumn{1}{l}{65.15} & \multicolumn{1}{l}{66.74} & \multicolumn{1}{l}{46.24} & \multicolumn{1}{l}{69.58} \\[-0.5ex]
                                                           & \multicolumn{1}{l}{iIoU} & \multicolumn{1}{l}{43.95} & \multicolumn{1}{l}{52.63} & \multicolumn{1}{l}{54.06} & \multicolumn{1}{l}{37.89} & \multicolumn{1}{l}{55.61}  \\[-0.5ex]
 \hline
\end{tabular}
}
\end{table*}
\begin{table*}
\centering
\caption{\small{ABLATION STUDY {\footnotesize{OF}} ASPN {\footnotesize{IN}} TERMS {\footnotesize{OF}} MEAN IOU PERCENTAGE, MEAN iIOU, {\footnotesize{AND}} COMPUTATION SPEED (FLOPs) {\footnotesize{ON}} EPFL DATASET. {\footnotesize{THE}} $\surd$ SHOWS {\footnotesize{THE}} PROPOSED NETWORK DIVISIONS THAT PARTICIPATED {\footnotesize{IN}} EACH PART {\footnotesize{OF}} EVALUATION.}}
  \label{tab:3}\small{
  \begin{tabular}{p{3.7cm} p{1cm}p{1cm}p{0.1cm}p{0.1cm}p{0.1cm}p{0.1cm}p{0.1cm}p{0.1cm}}     \hline                                                
\\ [-0.5em]
    Module & Channels & Base &   &   &   &   &   &   \\     \hline
\\ [-0.5em]
\multirow{2}{*}{+1~OUN} & 128 &  &\multicolumn{1}{l}{$\surd$} &   &   &  &  &    \\[-0.5ex]
                                            & 256 &  &   &   &  &   &   &    \\ \cline{1-2}  
\\ [-0.5em]
\multirow{2}{*}{+3~OUN} & 128 &  &  & \multicolumn{1}{l}{$\surd$} &   &   &   &    \\[-0.5ex]
                                            & 256 &  &  &   &   &   &   &    \\ \cline{1-2} 
\\ [-0.5em]
\multirow{2}{*}{+6~OUN} & 128 &  &  &  & \multicolumn{1}{l}{$\surd$}  &   & \multicolumn{1}{l}{$\surd$}  &  \multicolumn{1}{l}{$\surd$}  \\[-0.5ex]
                                            & 256 &  &  &   &   &   &   &    \\ \cline{1-2}
\\ [-0.5em]
\multirow{2}{*}{+8~OUN} & 128 &  &  &  &  & \multicolumn{1}{l}{$\surd$}   &    &     \\[-0.5ex]
                                           & 256 &  &  &   &   &   &   &    \\ \cline{1-2} 
\\ [-0.5em]
+SFC  &  &  &  &  &  &   &   $\surd$  &   $\surd$   \\  \cline{1-2} 
\\ [-0.5em]
VGGNet-16 to ResNet-101 &  &  &  &  &  &   &   &  $\surd$   \\  \hline 
\multirow{2}{*}{Mean IoU (\%)} & & \multirow{2}{*}{59.9} & \multicolumn{1}{l}{71.81} & \multicolumn{1}{l}{74.52} & \multicolumn{1}{l}{76.33} & \multicolumn{1}{l}{76.32} & \multicolumn{1}{l}{78.26} &	\multicolumn{1}{l}{\bf 81.52}  \\[-0.2ex]
                                            &  & & 72.57 & 75.28 & 76.95 & 76.88 & 79.51 & {\bf 81.68}   \\  \hline
\multirow{2}{*}{Mean IoU (\%)} & & \multirow{2}{*}{48.4} & \multicolumn{1}{l}{60.63} & \multicolumn{1}{l}{62.76} & \multicolumn{1}{l}{65.29} & \multicolumn{1}{l}{65.28} & \multicolumn{1}{l}{66.47} & \multicolumn{1}{l}{\bf 67.29}   \\[-0.2ex]
                                            &  & & 61.07 & 63.13 & 66.48 & 66.49 & 67.08 & {\bf 67.43}   \\  \hline
\multirow{2}{*}{FLOPs (B)}  &  & \multirow{2}{*}{30.67} &	52.24 & 61.86 & 68.27 & 70.95 & 72.71 & 80.78  \\[-0.2ex]
                                         &  &  & 58.79 & 70.27 & 74.83 & 82.32 & 83.2 & 84.42       \\ 
\hline
\end{tabular}}
\end{table*}
\begin{figure*}
  \centering
  \includegraphics[width=0.75\textwidth]{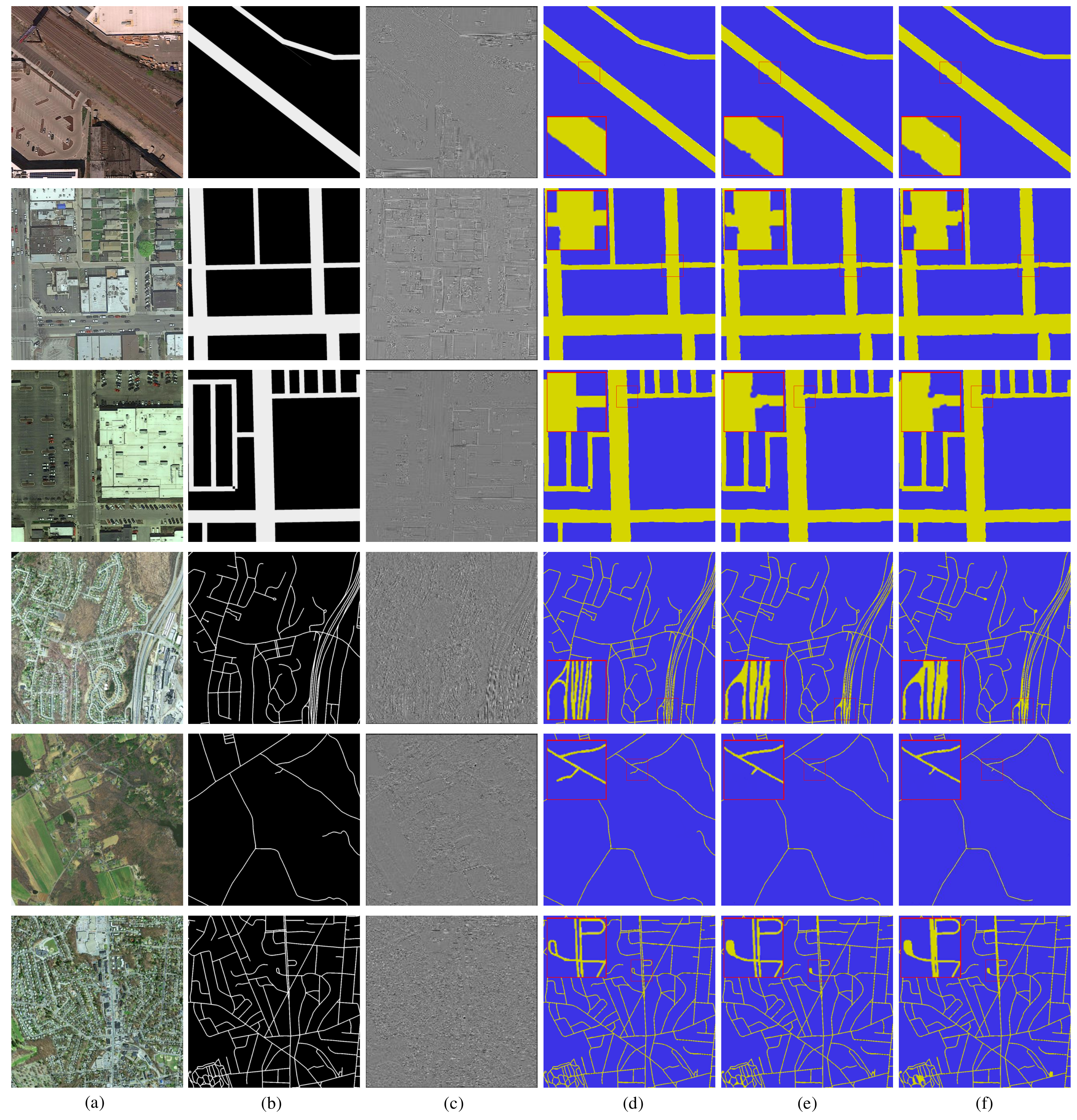}
\caption{\small Qualitative Results. (a). Input images. (b). Ground truth annotations. (c) Boundary detection by ASPN (ours). (d) Testing results of ASPN (ours). (e) Testing results of Hong et al.~\cite{hong2018conditional}. (f) Testing results of Sun and Wu \cite{sun2019learning}.  The red rectangles regions are close-ups for better visualisation.}
\label{fig:5}
\vspace{-1cm}
\end{figure*}
In Table~\ref{tab:1}, we compare the performance of ASPN with that of the other approaches; we also evaluate the performance of our model with and without the proposed pyramid network. In Table~\ref{tab:1}, we discuss the following aspects: The backbone network types, the initial input size to the network, the model strategy and the test results of the models. We also report the performance of the proposed pyramid network with different settings on the Massachusetts road dataset. It is remarkable that the proposed pyramid network with VGG-16 backbone has the best performance, which has outperformed the other state-of-the-art models even with more powerful backbones. IoU of Huang et al.~\cite{huang2013unified} is $25.6$, IoU of Zhu et al.~\cite{zhu2017unpaired} with FCN is $41.7$, IoU of Shrivastava et al.~\cite{shrivastava2017learning} with ResNet-101 is $44.5$ and IoU of Hong et al.~\cite{hong2018conditional} with VGG-19 is $45.5$. The performance of the proposed ASPN model is evaluated based on different network settings (four, six and eight OUNs while having ResNet-101 or VGG-16 as the backbone network) for two types of input size ($512\times512$ and $320\times320$). The IoU of ASPN that is assembled with FCN on $512$ and $320$ input image size is $44.8$ and $44.6$ respectively. As the results show, the ASPN with six OUNs, and $256$ channels has the best performance among the other implementations. Meanwhile, the performance of the ASPN with eight OUNs, is almost equal to that of six OUNs but the computation costs are higher. Additionally, in ASPN, both VGG-16 and ResNet-101 have better performance on the bigger size's input images and the best performance is achieved from the ASPN with VGG-16 on the image size of $512\times512$.
\begin{figure*}
\vspace{-1cm}
  \centering
  \includegraphics[width=0.75\textwidth]{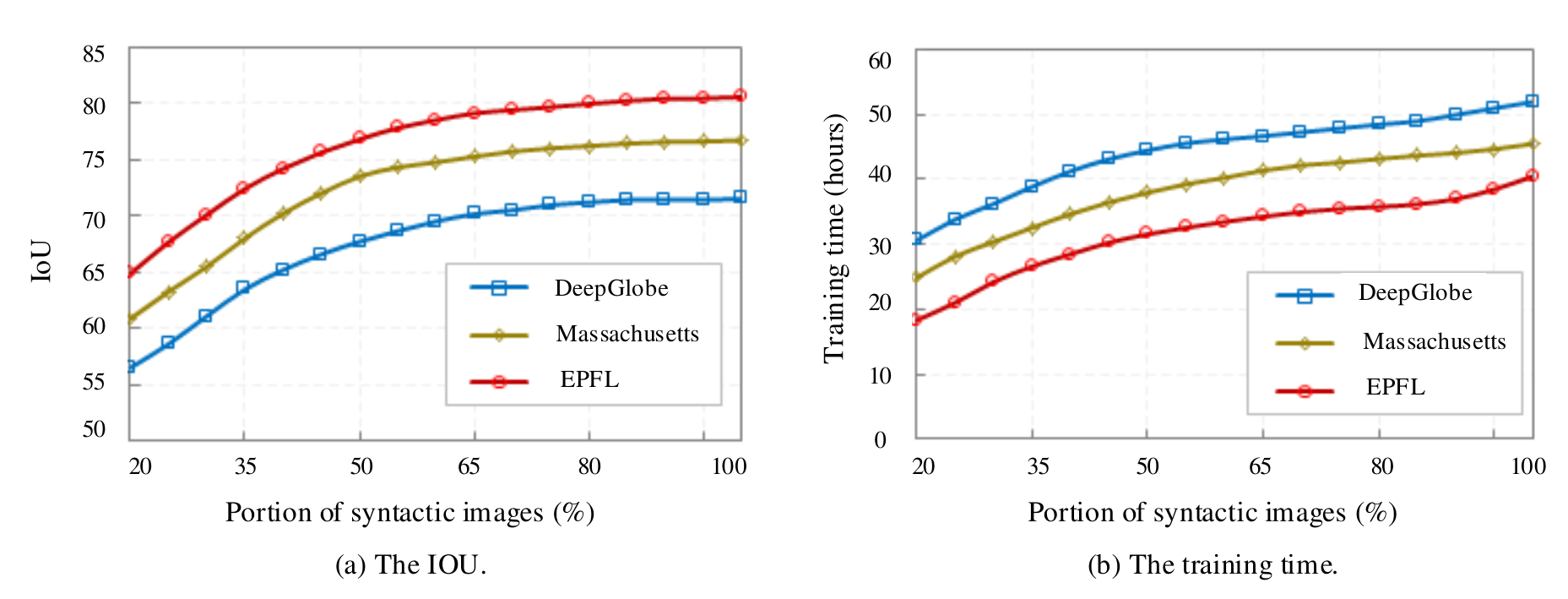}
\caption{\small (a)~The change of portions of synthetic images versus IoU. (b)~The change of portions of synthetic images versus training time.}
\label{fig:6}
\end{figure*}
\subsection{Analysis of Receptive field in Discriminator}
In Fig.~\ref{fig:4}, we showed some synthesis samples produced by the ASPN, plus the visualization of attention maps after fusing multiscale feature maps. In fact, Fig.~\ref{fig:4} shows the effect of fusing multilevel and multiscale features in synthesis image generation. In the discriminator, similar to \cite{hoffman2018cycada}, a $70\times 70$ receptive field is used to examine each structure at different scales. The discriminator uses a random Markov theory for classification. In this approach, even if a single $N\times N$ patch in an image is treated as a fake, then the classifier counts the image as a fake one. The added convolution layers, before fully connected layers in the discriminator, significantly increase the size of the receptive field without increasing the model depth and adding new parameters.
\subsection{Ablation study}
In this section, we evaluate the effects of different modules in the performance of the proposed ASPN model. The results of ASPN and the other state-of-the-art methods are reported in Tables~\ref{tab:1} and \ref{tab:2}. It has been observed, the domain adaptation models perform better that the other approaches. ASPN on all three datasets has higher IoU and iIoU as compared to others. The IoU of ASPN, minimum $3.8\%$ is higher than the other baselines and clearly justifies the proficiency of the proposed method.  
In comparison with the other approaches \cite{hong2018conditional, sun2019learning, hoffman2018cycada, shrivastava2017learning}, ASPN outperforms them by a gap of $3.5\% \sim 5.3\%$ in IoU. From Table~\ref{tab:2} and Fig.~\ref{fig:5}, we notice that the adaptation from EFPL and Massachusetts are higher than that of DeepGlobe, which is mainly due to a higher number of training images and the structure of the images. Moreover, we show the trend between the segmentation results and the amount of the synthetic data in Fig.~\ref{fig:6}.

Fig.~\ref{fig:6}(a) shows the IoUs of the proposed model on all three datasets. It demonstrates that: 1) Adopting more synthetic images can significantly improve the performance. For example, $30\%$ increase in the data, can improve the IoU by $9.5\%$. The best performance of our model is made on the EFPL and Massachusetts datasets respectively as the label maps in these datasets are generated by rasterizing road centrelines, and the average line thickness is about $15-20$ pixels with no smoothing. On the other hand, the DeepGlobe dataset is an edge-based dataset which was annotated based on the width of the road on the images. In this dataset, the average width of the road label is about $10$ pixels. Therefore, due to slim roads on the images in the DeepGlobe dataset, our model does not have high performance on this dataset. 2) In addition to the quantity of the synthetic data, the variety of images is also quite important for training a good model. Fig.~\ref{fig:6}(b) shows the training time of the ASPN with different amounts of the synthetic data. 

{\it 1)	The influence of the conditional generator} \\
\noindent To analyze the contribution of the proposed conditional generator towards quality improvement, we have carried out an evaluation for the models that were proposed by Xu and Zhao \cite{xu2018satellite}, Hoffman et al.~\cite{hoffman2018cycada} and two variations of the proposed network (trained with and without the conditional generator). In this evaluation, the conditional generator is removed and we only keep the discriminator, the feature extractor and the pixel-wise classifier. The results are listed in Table~\ref{tab:2}. The conditional generator performs better than other generators over all the three datasets. The results prove the capability of the proposed model in extracting multiscale features from the shallow and deep layers, resulting in high segmentation performance. By removing the conditional generator, both the IoU and iIoU drop around $25\%$. \\
\begin{figure*}
  \centering
  \includegraphics[width=0.75\textwidth]{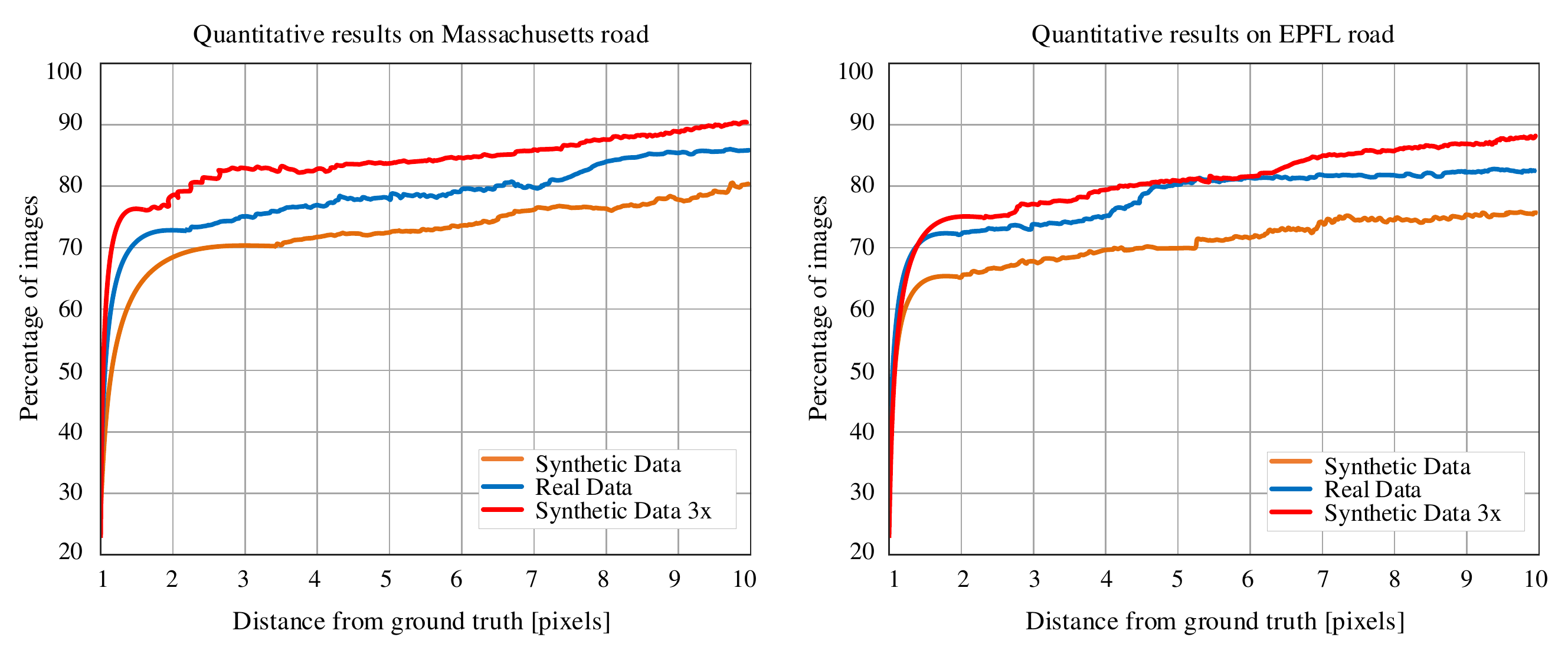}
\caption{\small Quantitative results for road segmentation on the test set of real images from Massachusetts and EPFL dataset. The plots represent cumulative curves as a function of distance from ground truth keypoint locations, for several numbers of training samples of synthetic images ($3\times$ denotes $100\%$ of the dataset).}
\label{fig:7}
\end{figure*}
{\it 2) The influence of different modules in learning} \\
As we have already discussed, ASPN is a collection of various subcomponents and here we analyze the effect of each module. \par
{\bf OUN:} to validate the effect of this module, three sets of experiments are conducted. Firstly, the base network is extended with a sequence of deconvolution layers. As a result, the IoU has improved from $59.9\%$ to $62.5\%$. Secondly, we used the combination of $3$ and $6$ OUNs in the generator and the performance has been improved to $74.52\%$ and $76.33\%$ on $128$ channels respectively. Finally, we used the stack of $8$ OUNs and the best performance reached to $76.32\%$ on $128$ channels.
 To figure out the best configuration for the proposed model regarding the number of OUNs and internal channels, we have conducted several trials. In these implementations, we use the EPFL dataset, the backbone network is VGG-16 and the input image size is $320\times320$. The details are listed in Table~\ref{tab:3}. From the experiments, it has been observed, increasing the number of OUNs and channels can improve the performance of the proposed model. However, it is worth to mention that, increasing the number of OUNs make it more efficient than increasing the number of internal channels. Even, by adding more number of OUNs, the number of the parameters almost remains constant.\par
{\bf SFC:} As shown in the $8^{th}$ column of Table~\ref{tab:3}, by using SFC modules, all the multiscale features are properly concatenated, which greatly helps to improve the performance.\par
{\bf Backbone network:} from the above tasks, it has been observed, replacing VGGNet-16 with ResNet-101 can slightly improve the performance of the proposed model. As shown in Table~\ref{tab:3}, the IoU goes from $81.52\%$ to $81.68\%$ by adopting the ResNet-101 as the base network. However, it increases the computation cost. \par
\begin{figure*}
  \centering
  \includegraphics[width=0.75\textwidth]{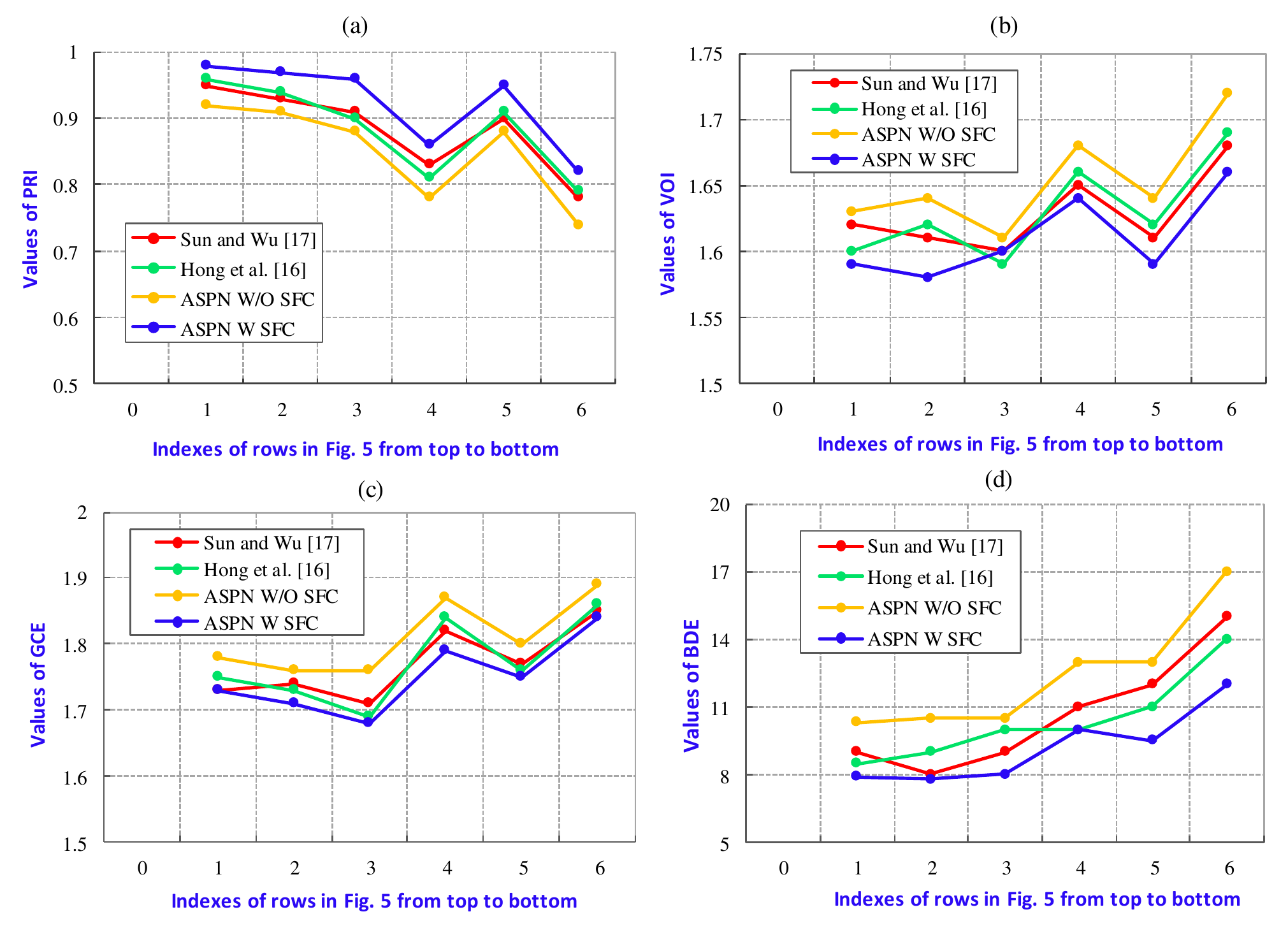}
\caption{\small The quantitative comparisons of the above mentioned two approaches and the proposed ASPN with and without the SFC module on synthetic images with the indexes of rows (from top to bottom in Fig.~\ref{fig:5}) as the horizontal axis. (a) PRI, (b) VOI, (c) GCE, (d) BDE.}
\label{fig:8}
\end{figure*}

{\bf Quantitative results:} We used our previous work \cite{shamsolmoali2019novel} to extract the roads from the synthetic samples. Fig.~\ref{fig:7} shows the performance of road segmentation model \cite{shamsolmoali2019novel} that is trained on the synthetic data and the real data. As the results show, increasing the number of the training data can improve the results. In this evaluation, the model is trained on the real data, half of the synthesis data and finally, all of the synthesis data. By using the entire synthesis data, there is around $11.8\%$ performance improvement. Fig.~\ref{fig:8} shows the quantitative comparisons of the PRI \cite{unnikrishnan2007toward}, VoI \cite{meila2005comparing}, GCE \cite{martin2001database} and BDE \cite{freixenet2002yet} for the four comparision approaches with the indexes of rows (from top to bottom in Fig.~\ref{fig:5}) as the horizontal axis. From the four charts of Fig.~\ref{fig:8}, it can be observed the values of the four performance measures for our proposed ASPN model with the SFC module are better than the other two approaches and our model without the SFC module, are slightly worse than the other approaches with multiscale feature network.\par
\begin{figure*}
  \centering
  \includegraphics[width=0.8\textwidth]{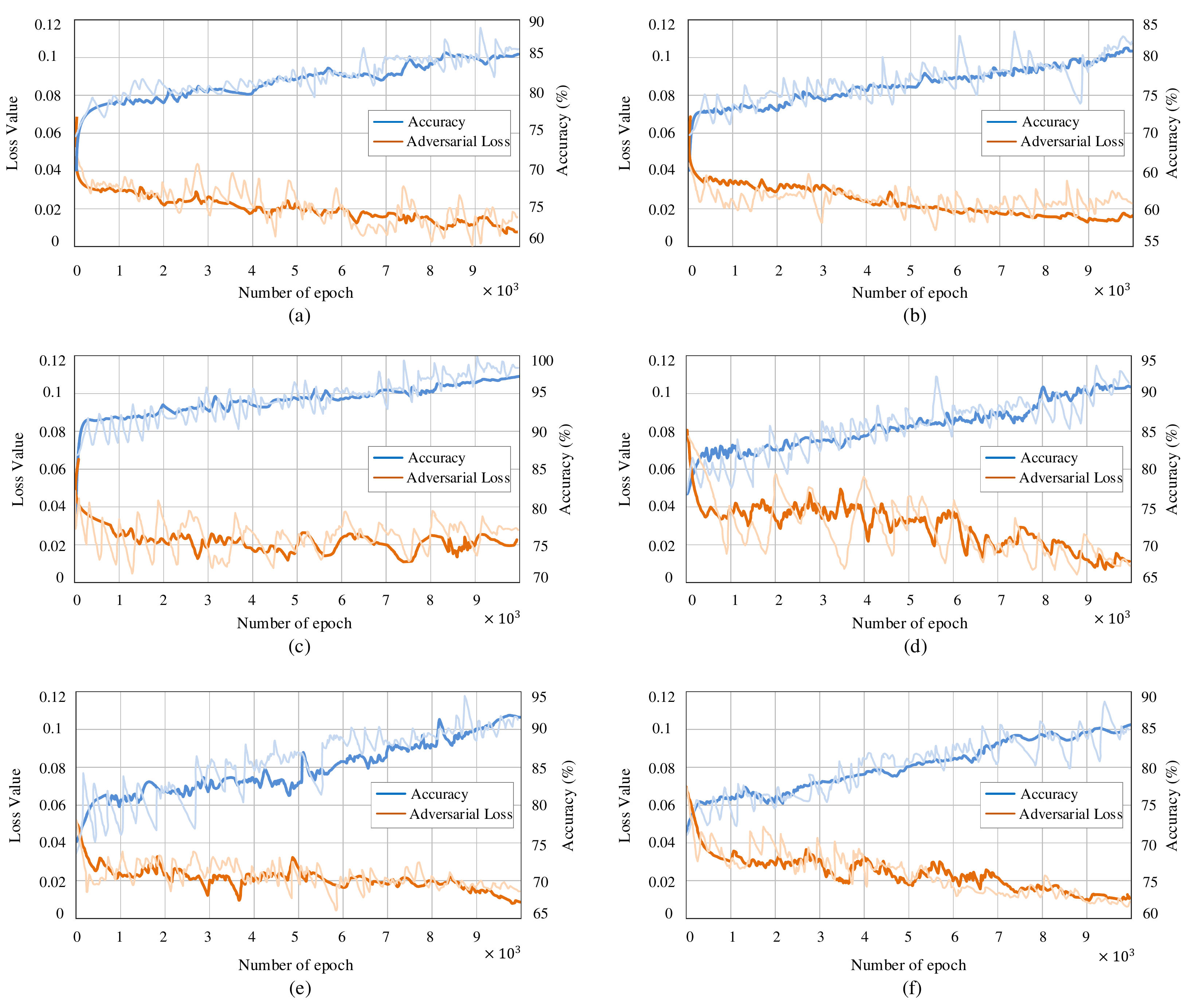}
\caption{\small Relationship between accuracy and adversarial loss while applying cross-domain training on EPFL $\leftrightarrow$  Massachusetts, EFPL $\leftrightarrow$  DeepGlobe, and Massachusetts $\leftrightarrow$  DeepGlobe datasets. (a) EPFL $\leftrightarrow$  Massachusetts. (b) EPFL $\leftrightarrow$  DeepGlobe. (c) Massachusetts $\leftrightarrow$  EPFL. (d) Massachusetts $\leftrightarrow$  DeepGlobe. (e) DeepGlobe $\leftrightarrow$  EPFL. (f) DeepGlobe $\leftrightarrow$  Massachusetts.}
\label{fig:9}
\vspace{-1cm}
\end{figure*}

{\bf Domain adaptation:} Fig.~\ref{fig:9} demonstrates the relation between the accuracy and adversarial loss during the cross-domain training of the six datasets. EFPL $\leftrightarrow$ Massachusetts, EPFL $\leftrightarrow$ DeepGlobe, and Massachusetts $\leftrightarrow$ DeepGlobe. The results indicate that, applying cross-domain training helps to minimize the adversarial loss and also increases the adaptation accuracy. In this evaluation, the best results are achieved from Massachusetts $\rightarrow$ EPFL, because, the number of the training images in the Massachusetts dataset is more than the other datasets and have a lower complexity. The second-best results are obtained from DeepGlobe $\rightarrow$ EPFL. 
\begin{figure*}
  \centering
  \includegraphics[width=0.75\textwidth]{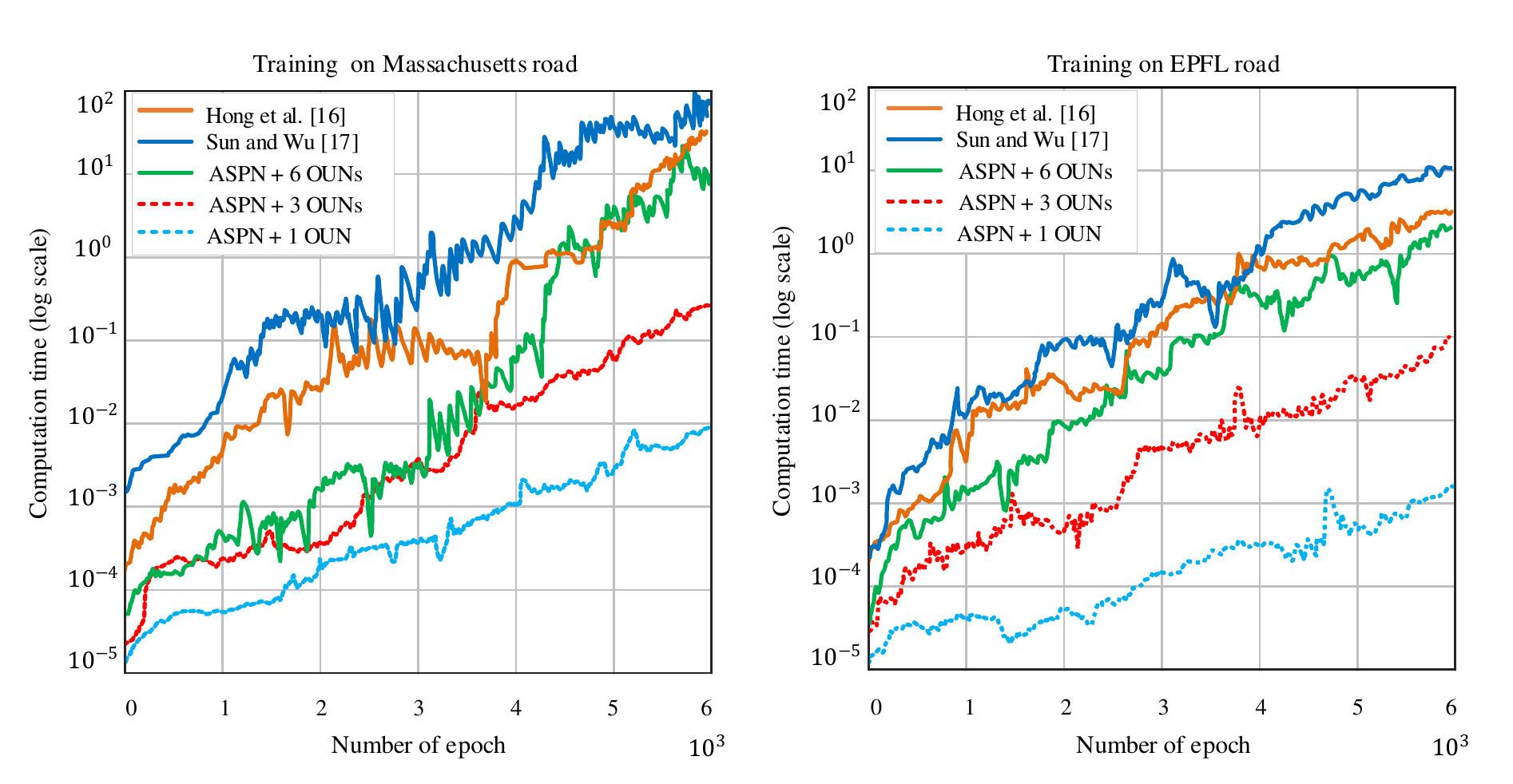}
 \caption{\small Computation time {\tt v.s.} a number of iterations while training on Massachusetts and EPFL dataset. The light blue, red and green refer to propose model with $1, 3$ and $6$ OUNs respectively. The orange shows the performance of Sun and Wu \cite{sun2019learning} and dark blue represents the performance of Hong et al.~\cite{hong2018conditional}.}
\label{fig:10}
\end{figure*}
\begin{table}
\centering
  \caption{\small{PERFORMANCE COMPARISON {\footnotesize{OF}} SEGMENTATION METHODS {\footnotesize{ON}} {\footnotesize{THE}} MASSACHUSETTS DATASET. {\footnotesize{THE}} BEST TWO RESULTS ARE HIGHLIGHTED {\footnotesize{IN}} \textcolor{ForestGreen}{GREEN} (BEST) {\footnotesize{AND}} \textcolor{red}{RED} (SECOND BEST)}}
  \label{tab:4}
  \begin{tabular}{lcccc}     \hline
\\ [-0.5em]
    Methods & PRI & VOI & GCE & BDE  \\     \hline
\\ [-0.5em]
Henry et al.~\cite{henry2018road} &	0.8227&	1.7048&	0.1938&	17.37    \\[-0.3ex]
Li et al.~\cite{li2018road} &   0.8288&        1.6911&        0.1851&       12.69   \\[-0.3ex]
Yu et al.~\cite{yu2018semantic} &   0.8329	&1.6885	&0.1837	&12.07   \\[-0.3ex]
Zhang et al.~\cite{zhang2019pan} &	0.8336	&1.6862	&0.1822	&11.68    \\[-0.3ex]
Sun et al.~\cite{sun2019learning} &	0.8352	&1.6853	&0.1810	&11.34   \\[-0.3ex]
Hong et al.~\cite{hong2018conditional} &    0.8356          &1.6851        &0.1804        &11.13   \\[-0.3ex]
Li et al.~\cite{li2019road} &            0.8358          &1.6845        &0.1798        &11.06    \\[-0.3ex]
Liu et al.~\cite{liu2018roadnet} & 	0.8359         &1.6839         &0.1795        & $\textcolor{red}{10.58}$   \\[-0.3ex]
Wei et al.~\cite{wei2020simultaneous} &       0.8360         & $\textcolor{red}{1.6836}$ & 0.1791	& $\textcolor{ForestGreen}{10.56}$  \\[-0.3ex]
Li et al.~\cite{li2019nested} &   $\textcolor{red}{0.8362}$	& 1.6844	&  $\textcolor{red}{0.1789}$	& 10.62   \\[-0.3ex]
ASPN& $\textcolor{ForestGreen}{0.8367}$	& $\textcolor{ForestGreen}{1.6833}$	& $\textcolor{ForestGreen}{0.1784}$	& $\textcolor{red}{10.58}$ \\[-0.3ex]
        \hline
\end{tabular}
\end{table}
\begin{table}
\centering
  \caption{\small{PERFORMANCE COMPARISON {\footnotesize{OF}} SEGMENTATION METHODS {\footnotesize{ON}} {\footnotesize{THE}} EPFL DATASET. {\footnotesize{THE}} BEST TWO RESULTS ARE HIGHLIGHTED {\footnotesize{IN}} \textcolor{ForestGreen}{GREEN} (BEST) {\footnotesize{AND}} \textcolor{red}{RED} (SECOND BEST)}}
  \label{tab:5}
  \begin{tabular}{lcccc}     \hline
\\ [-0.5em]
    Methods & PRI & VOI & GCE & BDE  \\     \hline
\\ [-0.5em]
Henry et al.~\cite{henry2018road} & 0.8638 	&1.6933	&0.1863	&17.20   \\[-0.3ex]
Li et al.~\cite{li2018road} & 0.8747	&1.6818	&0.1748	&12.45  \\[-0.3ex]
Yu et al.~\cite{yu2018semantic} &   0.8815	&1.6789	&0.1726	&11.87  \\[-0.3ex]
Zhang et al.~\cite{zhang2019pan} &	0.8826	&1.6781	&0.1693	&11.41   \\[-0.3ex]
Sun et al.~\cite{sun2019learning} &	0.8830	&1.6773	&0.1678	&10.86  \\[-0.3ex]
Hong et al.~\cite{hong2018conditional} &   0.8834           &1.6773        &0.1675        &10.74   \\[-0.3ex]
Li et al.~\cite{li2019road} &            0.8849          &1.6772        &0.1672       &10.66  \\[-0.3ex]
Liu et al.~\cite{liu2018roadnet} & 	0.8848         & $\textcolor{red}{1.6762}$       &0.1667        & $\textcolor{red}{10.45}$  \\[-0.3ex]
Wei et al.~\cite{wei2020simultaneous} &       0.8850        & $\textcolor{ForestGreen}{1.6760}$ & 0.1791	& $\textcolor{ForestGreen}{10.56}$ \\[-0.3ex]
Li et al.~\cite{li2019nested} &   $\textcolor{red}{0.8851}$	& 1.6770	&  $\textcolor{red}{0.1667}$	& 10.48  \\[-0.3ex]
ASPN& $\textcolor{ForestGreen}{0.8854}$	& $\textcolor{red}{1.6762}$	& $\textcolor{ForestGreen}{0.1663}$	& $\textcolor{ForestGreen}{10.43}$  \\[-0.3ex]
        \hline
\end{tabular}
\end{table}

Fig.~\ref{fig:10} shows the computation cost of the proposed ASPN model against the number of iterations based on the different implementation settings. As the results show, while the number of OUNs increases, the model demands more computational resources. However, still, the proposed model is more proficient than the state-of-the-art approaches. We compare the performance of ASPN with several methods on the Massachusetts road dataset in Table~\ref{tab:4} and the EPFL dataset in Table~\ref{tab:5}. Four performance metrics are used for quantitative evaluation (PRI \cite{unnikrishnan2007toward}, VoI \cite{meila2005comparing}, GCE \cite{martin2001database}, and BDE \cite{freixenet2002yet}) between two segmented images (the best and the second best results are highlighted in green and red, respectively).
The following conclusions have been observed from these tables. First, ASPN almost achieves the best performance in terms of all the four metrics. Second, Liu et al.~\cite{liu2018roadnet}, Wei et al.~\cite{wei2020simultaneous} and Li et al.~\cite{li2019nested} have better performance in the GCE and BDE and show that these three models have better boundary segmentation performance in comparison with the other models. In overall, ASPN shows excellent segmentation performance with better boundary adherence and less displacement errors with respect to the ground truth.

\section{Conclusion}
This paper has presented a novel architectural model for road segmentation on remote sensing images by using the adversarial networks and domain adaptation. A conditional generator based on feature pyramid networks was introduced, which has a wide structure to extract multilevel and multiscale features. The proposed model has several subcomponents that systematically extract the shallow and deep features on different scales, and gather various contextual information for generating the final feature maps. Hence, the model can preserve the structure of the elements in an image. The proposed model has superior performance in comparison with the other state-of-the-art approaches that use adversarial segmentation and domain adaptation. In the future, we plan to build a more efficient model to improve the segmentation accuracy not only on road but also on the other segmentation tasks whilst reducing the overall computation costs.

\footnotesize{
\bibliographystyle{IEEEtran}
\bibliography{IEEEabrv,IEEEexample}
}


\end{document}